\documentclass[10pt,twocolumn,letterpaper]{article}
\usepackage[pagenumbers]{cvpr} 
\usepackage[dvipsnames]{xcolor}
\usepackage{amsfonts}
\usepackage{bm}
\usepackage{amsmath}
\usepackage{mathtools}
\usepackage{multirow}
\usepackage{booktabs}
\usepackage{caption}
\usepackage{enumitem}

\definecolor{mygray}{RGB}{120, 120, 120}
\definecolor{myblue}{RGB}{68, 114, 196}
\definecolor{myorange}{RGB}{237, 125, 49}

\usepackage[accsupp]{axessibility}
\usepackage{graphicx}
\usepackage{amssymb}
\usepackage{booktabs}
\usepackage{amsmath}
\usepackage{colortbl}
\usepackage{color}
\usepackage{multirow}
\usepackage[dvipsnames]{xcolor}
\usepackage{mathtools}
\usepackage{gensymb}

\usepackage{stmaryrd}
\usepackage{trimclip}

\makeatletter
\DeclareRobustCommand{\shortto}{%
  \mathrel{\mathpalette\short@to\relax}%
}

\newcommand{\short@to}[2]{%
  \mkern2mu
  \clipbox{{.5\width} 0 0 0}{$\m@th#1\vphantom{+}{\shortrightarrow}$}%
  }
\makeatother

\usepackage{pifont}
\DeclareMathOperator*{\argmax}{arg\,max}

\usepackage[pagebackref,breaklinks,colorlinks]{hyperref}
 
\newcommand{\x}{\mathbf{x}}
\newcommand{\dir}{\mathbf{d}}
\newcommand{\origin}{\mathbf{o}}

\newcommand{\real}{\mathbb{R}}
\newcommand{\ray}{\mathbf{r}}
\newcommand{\density}{\sigma}
\newcommand{\opacity}{\alpha}
\newcommand{\reflectance}{\rho}
\newcommand{\radiance}{\mathbf{c}}
\newcommand{\intensity}{e}
\newcommand{\pdrop}{p_d}
\newcommand{\ptwo}{p_s}
\newcommand{\posfeat}{\mathbf{f}_{\text{pos}}}
\newcommand{\geofeat}{\mathbf{f}_{\text{geo}}}
\newcommand{\geofeatbar}{\bar{\mathbf{f}}_{\text{geo}}}
\newcommand{\dirfeat}{\mathbf{f}_{\text{dir}}}
\newcommand{\rangefeat}{\mathbf{f}_{\text{range}}}
\newcommand{\rayfeat}{\mathbf{f}_{\text{beam}}}

\definecolor{sem0}{rgb}{0.98431373, 0.70588235, 0.68235294}
\definecolor{sem1}{rgb}{0.70196078,0.80392157, 0.89019608}
\definecolor{sem2}{rgb}{0.8, 0.92156863, 0.77254902}
\definecolor{sem3}{rgb}{0.87058824, 0.79607843, 0.89411765}
\definecolor{ourgray}{rgb}{0.78, 0.78, 0.78}
\definecolor{error}{rgb}{0, 0.635, 1}
\definecolor{sdpoints}{rgb}{1.0, 0.706, 0.0}
\definecolor{hit}{rgb}{0.12156863,0.47058824,0.70588235}
\newcommand{\coolwarm}{\includegraphics[width=3em,height=0.8em]{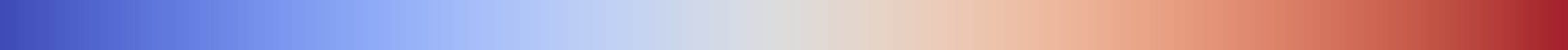}}
\newcommand{\bwr}{\includegraphics[width=3em,height=0.8em]{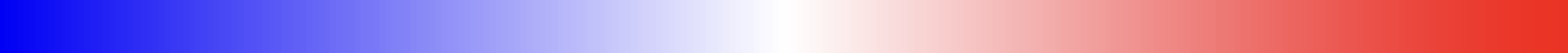}}

\DeclarePairedDelimiter\parens{\lparen}{\rparen}

\usepackage[capitalize]{cleveref}
\crefname{section}{Sec.}{Secs.}
\Crefname{section}{Section}{Sections}
\Crefname{table}{Table}{Tables}
\crefname{table}{Tab.}{Tabs.}
\crefname{equation}{\text{Eq}}{\text{Eq}}
\crefname{equation}{Eq.}{Eq.}

\definecolor{my_orange}{rgb}{1.0, 0.6, 0.0}

\def\cvprPaperID{1719} %
\def\confName{ICCV}
\def\confYear{2023}

\begin{document}

\title{Neural LiDAR Fields for Novel View Synthesis}

\author{
Shengyu Huang$^{1,2}$ \quad Zan Gojcic$^{2}$ \quad Zian Wang$^{2,3,4}$\quad Francis Williams$^{2}$ \\
Yoni Kasten$^{2}$ \quad Sanja Fidler$^{2,3,4}$ \quad Konrad Schindler$^{1}$ \quad Or Litany$^{2}$ \\
{\small $^{1}$ ETH Zurich \quad $^{2}$ NVIDIA \quad $^{3}$ University of Toronto \quad $^{4}$ Vector Institute
}
\\
\small \url{https://research.nvidia.com/labs/toronto-ai/nfl/}
}

\maketitle

\begin{abstract}
We present Neural Fields for LiDAR (NFL), a method to optimise a neural field scene representation from LiDAR measurements, with the goal of synthesizing realistic LiDAR scans from novel viewpoints. 
NFL combines the rendering power of neural fields with a detailed, physically motivated model of the LiDAR sensing process, thus enabling it to accurately reproduce key sensor behaviors like beam divergence, secondary returns, and ray dropping.
We evaluate NFL on synthetic and real LiDAR scans and show that it outperforms explicit reconstruct-then-simulate methods as well as other NeRF-style methods on LiDAR novel view synthesis task. Moreover, we show that the improved realism of the synthesized views narrows the domain gap to real scans and translates to better registration and semantic segmentation performance.
\end{abstract}
\section{Introduction}
\label{sec:intro}

The goal of novel view synthesis is to generate a view of a 3D scene, from a viewpoint at which no real sensor image has been captured. This offers the possibility to observe \emph{real} scenes from a \emph{virtual}, unobserved perspective. Among other applications, it has tremendous potential for autonomous driving: synthetic novel views may be used to train and test perception algorithms across a wider range of viewing conditions, thus enhancing robustness and generalization. Moreover, novel view synthesis becomes critical when the desired viewpoints are not known in advance, \eg, during training of a planning module whose decisions determine future vehicle locations.

Neural radiance fields (NeRFs) have led to unprecedented visual quality when synthesizing novel camera views~\cite{mildenhall2020nerf, barron2021mip, yu2021plenoxels, muller2022instant}. These methods represent the 3D scene in form of continuous density and radiance fields, from which images can be generated through volume rendering, mimicking the image acquisition process. The inductive bias of neural networks imparts NeRFs the ability to interpolate complex lighting and reflectance behaviours with a high degree of realism.

While most prior works focused on synthesizing camera views, 3D perception in the autonomous driving context typically relies partly (or even exclusively) on LiDAR measurements. Synthesizing realistic LiDAR scans from novel viewpoints thus has a lot of potential for data augmentation and closed-loop testing of autonomous navigation systems.

The problem of synthesizing novel LiDAR views has previously been addressed in two stages~\cite{manivasagam2020lidarsim}. First, extract an explicit surface representation such as surfels or a triangular mesh from the scanned point clouds. Then, simulate LiDAR measurements from a novel viewpoint by casting rays and intersecting them with the surface model.
Like for images, explicit reconstruction (which is not optimised towards the subsequent synthesis step) suffers from discretization artifacts and introduces noticeable errors~\cite{waechter2014let}. Moreover, the rendering assumes an idealised ray model and neglects the divergence of the LiDAR beams, which causes frequent second returns from distant surfaces. 

Here, we instead build on a main insight of NeRF~\cite{mildenhall2020nerf}: directly optimizing an implicit scene representation for novel view synthesis can produce more realistic outputs than the reconstruct-then-simulate approach. Specifically, we propose Neural Fields for LiDAR (NFL), a NeRF-style representation for synthesizing novel LiDAR viewpoints.

Several NeRF extensions have utilized range measurements as additional supervision, and have shown that constraining the scene geometry more tightly can yield better (camera) view synthesis~\cite{deng2021depth,rematas2021urban}. Yet, the output of those methods are synthetic images, not LiDAR scans, consequently they have not paid attention to effects specific to LiDAR sensing: a laser scanner does \emph{not} directly sense range, rather it measures the returned light energy per ray and determines the range based on the waveform. 
This includes the possibilities that there are multiple returns\footnote{In principle there can be \textgreater2 returns, but automotive LiDAR sensors typically record the first two echos.} from the emitted ray, or no return at all.

Our formulation closely adheres to the principles of the LiDAR measurement process 
and incorporates them into the neural field framework. Specifically, we (\romannumeral 1)~\textbf{devise volume rendering for LiDAR sensors};  
(\romannumeral 2)~\textbf{incorporate beam divergence} and (\romannumeral 3)~\textbf{propose truncated volume rendering} to account for secondary returns and improve range prediction. 

We evaluate our method on both synthetic and real LiDAR data. To this end, we (\romannumeral 4)~\textbf{develop a LiDAR simulator} for synthesizing scenes from 3D assets that serve as a test bed for viewpoints far from the original scan locations, and to study the effect of different scan patterns.
Real data from the Waymo~\cite{sun2020scalability} dataset is used to evaluate NFL against real scans at held-out viewpoints, including real-world intensities, ray drops and secondary returns. Additionally, we (\romannumeral 5)~\textbf{propose a novel closed-loop evaluation protocol} that leverages real data to evaluate view synthesis in challenging views.
As an end-to-end test for downstream tasks, we further evaluate the performance of state-of-the-art segmentation and registration networks when trained on real scans and tested on novel views generated by NFL.

\section{Related Work}
\label{sec:relwork}
\paragraph{LiDAR simulation.}
Simulating realistic LiDAR data is useful for training perception models. Different from real-world LiDAR data that requires annotation efforts, simulated data can be automatically generated with ground truth labels, \eg object bounding boxes and semantic segmentation. Unrealistic LiDAR simulation will prevent the trained models from generalizing to real data.   %
Traditional simulation engines, such as those proposed in~\cite{dosovitskiy2017carla,koenig2004design}, require the specification of sensor parameters and 3D scene assets and use ray-casting methods for simulation. 
 Although these point clouds can accurately represent scene geometry, they often exhibit a discrepancy, or "domain gap", compared to real data, due to the lack of modeling for sensor noise, such as ray drop and Gaussian beam. 
Furthermore, this approach relies heavily on the creation of 3D scene assets, which can be time-consuming and expensive. 
To address these challenges, LiDARsim~\cite{manivasagam2020lidarsim} reconstructs the static and dynamic scene assets from real data using surfel~\cite{pfister2000surfels} representation and models the ray-drop pattern for improved realism. BaiduSim~\cite{baidusim} proposes a probability map to model scene compositions in order to reduce the domain gap. Most recent work~\cite{guillard2022learning} learns to enhance existing simulated LiDAR intensity and ray-drop patterns, using the available corresponding RGB images. 

Weather conditions, such as fog or rain, can significantly impact the quality of LiDAR data, and downstream models trained solely on ideal weather conditions may fail to generalize to these effects. Recent methods and datasets \cite{hahner2021fog,hahner2022lidar,kilic2021lidar,bijelic2020seeing} have been proposed to address this issue. %
  SnowSIM~\cite{hahner2022lidar} and FogSIM~\cite{hahner2021fog} sample snow particles and model the impulse response from atmospheric attenuation, respectively, to alter the range measurements of each ray.
 Other approaches~\cite{shih2022reconstruction,kilic2021lidar, kurup2021dsor} simulate LiDAR data on rainy days and the spray effects, in a similar fashion.

\vspace{-3mm}
\paragraph{NeRF for Novel View Synthesis.}
NeRF~\cite{mildenhall2020nerf} maps 5D position and direction to density and radiance scene values, and uses volume rendering~\cite{max1995optical,max2005local} to estimate pixel color. This technique has proven effective for generating realistic images at unseen camera views. Many methods have been proposed to improve robustness to camera poses~\cite{lin2021barf,chng2022garf}, handle dynamics~\cite{ost2021neural,pumarola2021d}, anti-alias~\cite{zhang2020nerf++,barron2021mip,barron2022mip}, and speed up optimisation~\cite{liu2020neural,yu2021plenoxels,muller2022instant} \etc. Despite its high-quality novel view synthesis capacity, the underlying geometry of NeRF is considered inaccurate and noisy~\cite{oechsle2021unisurf}, making it less favoured for geometry reconstruction, especially in sparse-views settings. ~\cite{oechsle2021unisurf,yariv2021volume,wang2021neus} address this challenge by using implicit surface representations, and defining the density functions based on them to enabling volume rendering. DS-NeRF~\cite{deng2021depth} and DenseDS-NeRF~\cite{roessle2021dense} use sparse depth supervision from SfM~\cite{schoenberger2016sfm} points to regularise the density field. Urban Radiance Field~\cite{rematas2021urban} leverages LiDAR data for depth supervision. %

\vspace{-3mm}
\paragraph{Neural fields beyond regular cameras.}
Neural fields are a natural and continuous representation~\cite{xie2022neural} for spatio-temporal information including SDFs~\cite{park2019deepsdf}, occupancy~\cite{mescheder2019occupancy} and radiance field \cite{mildenhall2020nerf} \etc. 
While in its original form, NeRF~\cite{mildenhall2020nerf} performs novel view synthesis using tonemapped low dynamic range images, RawNeRF~\cite{mildenhall2022nerf} extends it to operate over the high dynamic range images, enabling additional adjustments to focus, exposure, and tonemapping. T{\"o}rf~\cite{attal2021torf} incorporates the image formation model for continuous-wave Time-of-Flight (ToF) cameras into NeRF, allowing it to jointly process RGB and ToF sensor data and improve reconstruction robustness to large motions. EventNeRF~\cite{rudnev2022eventnerf} and ENeRF~\cite{klenk2022nerf} optimise the scene representation for Novel View Synthesis (NVS) from sparse event streams that contain asynchronous per-pixel brightness change signals. Other works~\cite{qadri2022neural,luo2022learning} explore the use of acoustic signals for surface reconstruction or NVS. 
\begin{figure*}[t]
\centering
\includegraphics[width=1.0\textwidth]{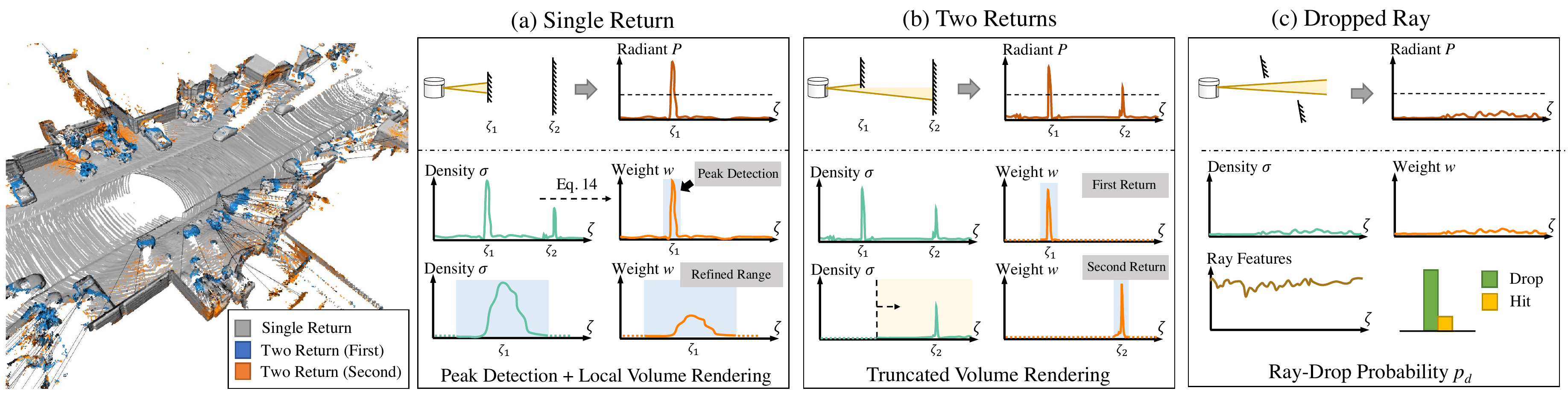}
\caption{Left: real LiDAR scan demonstrating key LiDAR return properties: a \textcolor{mygray}{single return} and two returns (first return shown in \textcolor{myblue}{blue} and second return in \textcolor{myorange}{orange}). Right: NFL models the waveform and accurately reproduces these properties. (a) Top: the LiDAR energy is fully scattered by the first surface. Bottom: NFL estimates range via peak detection on the computed weights $w$ followed by volume rendering based range refinement. (b) Top: secondary returns resulting from a beam hitting two surfaces. Bottom: NFL employs beam divergence and a truncated volume rendering to estimate the second return. (c) Top: beams that do not hit a surface do not return detectable signal. Bottom: NFL utilizes geometric and semantic features to predict the ray drop probability. Refer to section \ref{sec:render_lidar} for more details. }
\label{fig:overview}
\end{figure*}

\section{Background}
We start by reviewing the principles of volume rendering (\cref{sec:revisit_vr}) and the sensor model for LiDAR (\cref{sec:lidar_model}). This sets the stage for the proposed formulation of Neural LiDAR Fields~(\cref{sec:method}).

\subsection{Volume rendering for passive sensors}
\label{sec:revisit_vr}
In the following, we provide a brief summary of camera-based volume rendering as used by NeRF~\cite{mildenhall2020nerf,tagliasacchi2022volume}. This will serve as the basis to derive volume rendering equations for the active LiDAR sensor.
\vspace{-3mm}
\paragraph{Density and transmittance.}
For a ray $\ray(\origin, \dir)$ emitted from the origin $\origin \in \real^3$ in direction $\dir \in \real^3$, the \textit{density} $\density_\zeta$ at range $\zeta$ is a scalar function that indicates the differential likelihood of hitting a reflective particle at position $\ray_\zeta = \origin + \zeta \dir$. \textit{Transmittance} $T_{\zeta}$ indicates the probability of traversing the interval $[0, \zeta)$ without hitting anything. Taking a differential step $d\zeta$ along the ray,  the probability of \emph{not} hitting anything is $T_{\zeta+d\zeta} = T_{\zeta} \cdot \parens*{1 -\density_\zeta d\zeta}$.
Integrating over an interval $[\zeta_0, \zeta)$ yields the probability $T_{\zeta_0 \rightarrow \zeta}$ of traversing the interval unhindered,

\begin{equation}
T_{\zeta_0 \rightarrow \zeta} \equiv \frac{T_{\zeta}}{T_{\zeta_0}} = \exp\parens*{-\int_{\zeta_0}^\zeta \density_t dt}\;,
\label{eq:trans_ab}
\end{equation}
leading to the decomposition: $T_{\zeta} = T_{0 \rightarrow \zeta_0} \cdot T_{\zeta_0 \rightarrow \zeta}\;.$

\vspace{-3mm}
\paragraph{Integration over homogeneous media.}
Assuming a homogeneous medium along the ray segment $[\zeta_j, \zeta_{j+1}]$ with constant radiance $\radiance \in \real^3$ and density $\density$, the accumulated radiance from that segment evaluates to 
\begin{equation}
    \radiance(\zeta_j \!\shortto\! \zeta_{j+1}) 
    = \radiance_{\zeta_j}\int_{\zeta_j}^{\zeta_{j+1}} T_{\zeta_j \shortto \zeta} \cdot \density_\zeta \; d\zeta
    = \opacity_{\zeta_j} \radiance_{\zeta_j}\;,
\end{equation}
with $\opacity_{\zeta_j}=1-\exp\parens*{{-\density_{\zeta_j}(\zeta_{j+1} - \zeta_j)}}$ being the \textit{opacity}. 
\vspace{-3mm}
\paragraph{Volume rendering.}
By discretizing the ray into $N$ segments with piecewise constant densities and radiance values, we obtain the total irradiance (color to be rendered): %
\begin{equation}
    \begin{aligned}
      \radiance = \sum_{j=1}^N \int_{\zeta_j}^{\zeta_{j+1}}  T_{\zeta} \cdot \density_\zeta \radiance_\zeta \;d\zeta  
      = \sum_{j=1}^N  w_j \radiance_{\zeta_j}\;,
    \end{aligned}
    \label{eq: vol_render_nerf}
\end{equation}
where $w_j$ is the \textit{weight} for the $j$-th segment:
\begin{equation}
    \label{eq:weights_nerf}
    w_j = \opacity_{\zeta_j}\prod_{k=1}^{j-1}(1 - \opacity_{\zeta_k})\;.
\end{equation}
\subsection{LiDAR model}
\label{sec:lidar_model}

\begin{figure}[t]
\centering
\includegraphics[width=1.0\columnwidth]{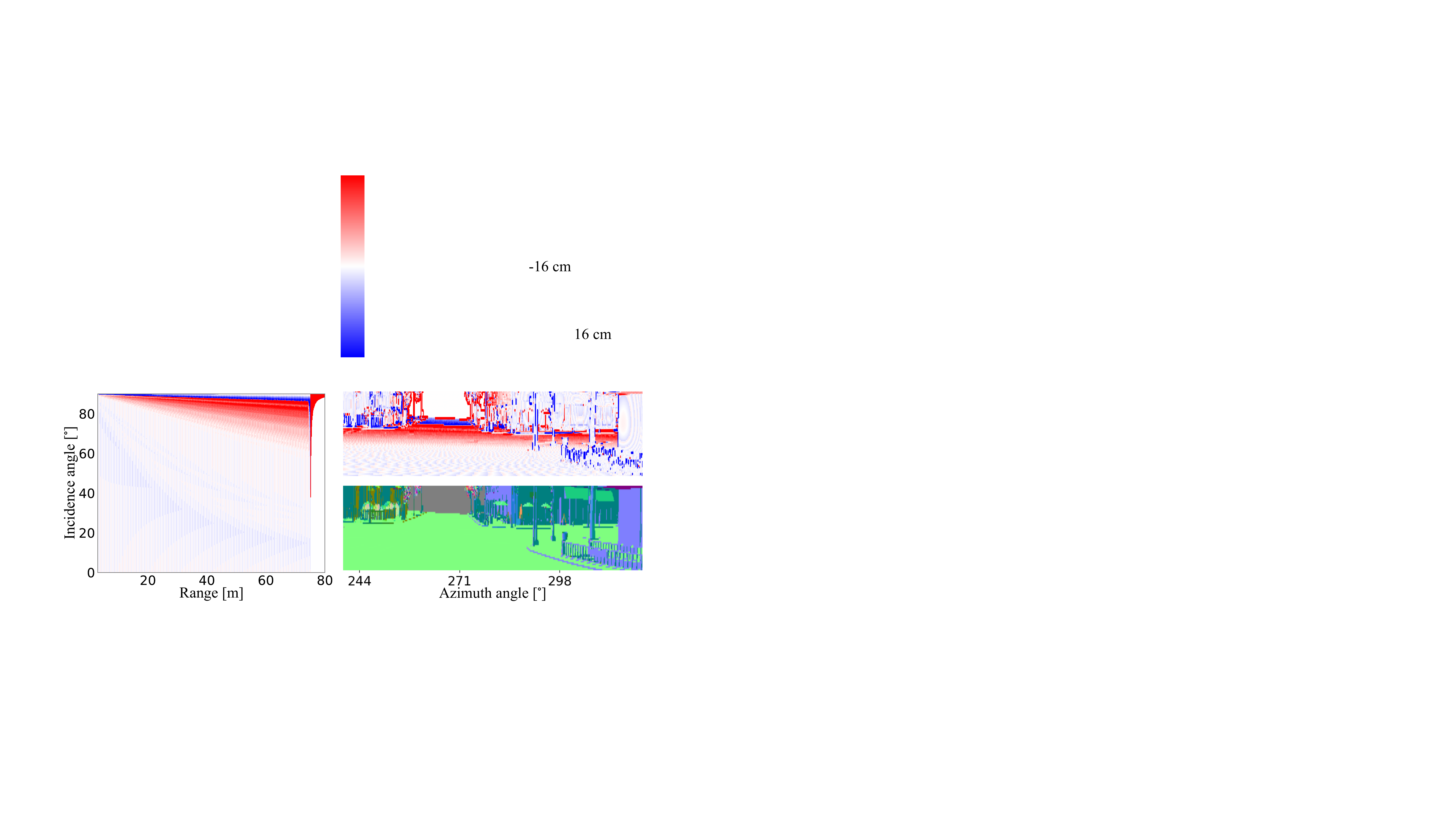}
\caption{The range accuracy of the LiDAR sensor is affected by waveform discretization and beam divergence. %
The LiDAR sensor has a tendency to overestimate range in high incidence angle regime, which becomes increasingly pronounced at higher range regimes (left). This is also reflected on \textit{TownReal} dataset (right).}
\label{fig:toy_example}
\end{figure}

LiDAR emits laser beam pulses and determines the distance from the sensor to the nearest reflective surface by measuring the time of flight. 
Often the LiDAR beams are pictured as ideal straight-line segments ending on a 3D surface point. In reality, things are more complicated: real lasers emit a pulse with non-zero divergence and finite pulse width, while real receivers employ signal processing techniques like radiant thresholding and binning to detect the return. This leads to phenomena such as discretization errors, over- and underestimation biases~(\cf \cref{fig:toy_example}), and multiple returns from one beam (or no return at all). 
In the following, we discuss key aspects of the LiDAR acquisition process and explain the effects that emerge, which inspire our model design. We also built a LiDAR simulator that accounts for these mechanisms, see~\cref{sec:dataset}.

\vspace{-3mm}
\paragraph{Beam divergence.}
LiDAR beams diverge as they travel away from the sensor. The size of laser beams can become wider over distance, and typically not negligible in street scenes. Consequently, the illuminated area grows and the irradiance (radiant power per area) decreases with increasing range.
The size of the beam's footprint is characterised by the divergence angle $(2\gamma_0)$ and the range $\zeta$. Let $\ray^{\gamma}$ be an ideal ray within the beam's cross-section, $\gamma \leq \gamma_0$, then its irradiance $E(\zeta, \gamma)$ at range $\zeta$ can be approximated by a Gaussian function in the ray coordinate system~\cite{wagner2006gaussian}:
\begin{equation}
    E(\zeta, \gamma) =\frac{2I_0}{\pi (\gamma_0 \zeta)^2} g(\gamma), \quad g(\gamma) = \exp\parens*{-2 \frac{\gamma^2}{\gamma_0^2}},
\end{equation}
where $I_0$ is the pulse peak power. %

\vspace{-3mm}
\paragraph{Pulse waveform.}
When the emitted LiDAR pulse returns to the sensor, the range to the reflective surface can be determined from its travel time and the speed of light $c$. Since the pulse has finite duration $\tau_H$, the time of return is found by analysing the received intensity profile. The transmitted pulse power over time can be characterised as~\cite{carlsson2001signature}:
\begin{equation}
    P_e(t) \propto \parens*{\frac{t}{\tau}}^2 \exp\parens*{-\frac{t}{\tau}},\quad \tau = \frac{\tau_H}{1.75}\;.
\label{eq:pulse}
\end{equation}
The range-dependent received radiant power $P(\zeta$) is the result of convolving the pulse power with the systems impulse response $H(\zeta)$~\cite{rasshofer2011influences,hahner2021fog,hahner2022lidar}:
\begin{equation}
    P(\zeta) = \int_0^{2\zeta/c} P_e(t) H(\zeta - \frac{ct}{2}) \; dt\;,
\end{equation}
where the impulse response $H(\zeta)$ is a composition of the target and the receiver responses: $H(\zeta) =  H_T(\zeta) H_C(\zeta)$.
Assuming a Lambertian surface, the target response due to a surface located at range $\zeta_0$ depends on the incidence angle $\theta$ and the reflectance $\reflectance$:
\begin{equation}
    H_T(\zeta) = \frac{\reflectance}{\pi} \cos(\theta) \delta(\zeta - \zeta_0)\;, 
\label{eq:ht}
\end{equation}
with $\delta(\cdot)$ the Dirac delta function.
The receiver response $H_C(\zeta)$ is computed by integrating over the solid angle spanned by the receiver's effective area $A_e$:
\begin{equation}
   H_C(\zeta) = T^2_{\zeta} \frac{A_e}{\zeta^2}\;,
\label{eq:hc}
\end{equation}
where $T_{\zeta} \in [0,1]$ is the one-way transmittance, squared to account for the two-way trip.

\vspace{-3mm}
\paragraph{Beam discretization.}
In practice, we follow~\cite{winiwarter2022virtual} and approximate the Gaussian beam profile using $M\!=\!37$ rays that are radially distributed around the central ray with different divergence angles $\gamma_i$. The total radiant power $P(\zeta)$ is the weighted sum over those rays: $P(\zeta) = \sum_{i=1}^M g(\gamma_i) P_i(\zeta)\;.$
Taking into account the beam divergence is important to reproduce two important phenomena: range biases and multiple returns, see \cref{fig:overview} and \cref{fig:toy_example}. As different rays hit a slanted surface at different ranges the integrated waveform peak may shift, causing over- or underestimations. Along object edges, rays within the same beam may hit different surfaces, causing multiple peaks (respectively, range readings), in the return waveform.

\vspace{-3mm}
\paragraph{Range estimation.}

One common approach to estimate the surface range from the received waveform is to locate its peak. To that end the signal is discretized in time to obtain a histogram, and local maxima above a certain threshold are declared detections~\cite{winiwarter2022virtual}. The associated range values are then corrected to remove known biases stemming from the pulse waveform (\cf \cref{eq:pulse}) and, optionally, biases due to the radiant power~\cite{winiwarter2022virtual}.
By modeling the binning and thresholding procedure one can reproduce further LiDAR  behaviors: systematic discretization errors in the range resolution (\cf \cref{fig:toy_example}), and the dropping of rays with low returned power (\cf \cref{fig:overview}).

\section{LiDAR Novel View Synthesis}
\label{sec:method}

We now turn to constructing a neural field model tailored for LiDAR scans, along with a differentiable volume rendering scheme to enable LiDAR novel view synthesis.
We first formulate the problem setting, then set up a corresponding neural scene representation (\cref{sec:neural_scene_rep}) and derive volume rendering for active sensing (\cref{sec:lidar_vr}). Finally, we describe the rendering procedure used to synthesize novel views (\cref{sec:render_lidar}) and our optimisation scheme (\cref{sec:opt}). 

\vspace{-3mm}
\paragraph{Problem setting.}
Consider a collection of LiDAR scans $\mathcal{X} = \{\mathbf{X}_v\}_{v=1}^{n_v}$ captured by a moving sensor (\eg, mounted on a vehicle). Each scan $\mathbf{X}_v$ is associated with a sensor pose $\mathbf{T}_v \in \text{SE}(3)$ and consists of $n_r$ rays. Every ray $\ray(\origin, \dir)$ records observations $(\zeta_1, \intensity_1, \pdrop, \ptwo, \zeta_2, \intensity_2)$: the range $\zeta_1$ and intensity $\intensity_1$ of the first return; a ray drop flag $\pdrop \in \{0, 1\}$; a two-return mask $\ptwo \in \{0,1\}$; and range $\zeta_2$ and intensity~$\intensity_2$ values of the second return.
Our goal is to reconstruct a (continuous) volumetric representation of the scene in terms of density $\density$ and reflectance $\reflectance$, from which we can subsequently render virtual LiDAR scans $\mathbf{X}_{tgt}$ from novel sensor poses $\mathbf{T}_{tgt}$.

\subsection{Neural scene representation}\label{sec:neural_scene_rep}
We encode the scene as a neural field $F: (\x, \dir) \mapsto (\density, \reflectance, \pdrop)$ that takes as input a location $\x \in \real^3$ and viewing direction $\dir \in \real^3$, and returns a density $\density$, a reflectance $\reflectance$ and a ray drop probability $\pdrop$. We found it beneficial to additionally return also a local contribution $\pdrop \in [0, 1]$ to the probability of ray drop, which will be discussed below.
Technically, we use a hash encoding~\cite{muller2022instant} to map coordinates $\x$ to positional features $\posfeat \in \real^{32}$ and project the view direction onto the first 16 coefficients of the spherical harmonics basis, $\dirfeat \in \real^{16}$. The neural field is parameterized by four Multi-Layer Perceptrons (MLPs): $[\density; \geofeat] = f_\density(\posfeat)$ regresses density and extracts an additional geometry feature $\geofeat \in \real^{15}$ that supports the other networks; $\reflectance = f_{\reflectance}(\geofeat, \dirfeat)$ regresses reflectance; $\pdrop = f_{\text{drop}}(\geofeat, \dirfeat)$ classifies whether a ray drop occurs; and  $p_s = f_{\text{sr}}(\rayfeat)$ classifies the existance of a second return. The feature $\rayfeat$ will be detailed in \cref{sec:render_lidar}.

\subsection{Volume rendering for LiDAR rays}
\label{sec:lidar_vr}
In contrast to passive sensors like cameras that rely on ambient illumination, LiDAR actively illuminates the scene and measures the back-scattered radiance. This two-way transmittance alters the volume rendering formulation. %

\paragraph{Radiant power integration.}
As discussed in \cref{sec:lidar_model} 
the radiant power along a LiDAR ray is a delta function that is non-zero only at reflecting surfaces. To incorporate this forward model into the volumetric representation we combine \cref{eq:ht} and \cref{eq:hc} to obtain the 
probabilistic radiant power:
\begin{equation}
P_\zeta = C \frac{T^2_{\zeta} \cdot \density_\zeta  \reflectance_\zeta}{\zeta^2} \cos(\theta)\;,
\label{eq:radiance}
\end{equation}
where $C$ is a system constant, $\reflectance_\zeta$ is the differentiable reflectance, and $\theta$ is the incidence angle.
In a homogeneous medium with constant reflectance $\reflectance$ and density $\density$, the integrated $P(\zeta_j \!\!\shortto\! \zeta_{j+1})$ evaluates to:
\begin{equation}
     P(\zeta_j\!\!\shortto\!\zeta_{j+1}) 
     \!=\!\!\!\int_{\zeta_j}^{\zeta_{j+1}}\!\!\!\!C \frac{T^2_{\zeta_j\!\shortto \zeta} \density_\zeta \reflectance_\zeta}{\zeta^2}\!\cos(\theta_{j})\,d\zeta
     \!\approx\!\opacity_{\zeta_j} \reflectance_{\zeta_j}', 
\label{eq:lidar_rad}
\end{equation}
where we approximate $\zeta \in [\zeta_j, \zeta_{j+1}]$ by $\frac{\zeta_j + \zeta_{j+1}}{2}$, and 
\begin{equation}
    \opacity_{\zeta_j}\!=\!\frac{1}{2}\parens*{1- e^{-2\density_{\zeta_j}\delta_j}}\;,\; \reflectance_{\zeta_j}' = C \reflectance_{\zeta_j}\frac{4 \cos(\theta_j)}{(\zeta_j + \zeta_{j+1})^2}.
\end{equation}

\vspace{-3mm}
\paragraph{Volume rendering.}
The observed power at the active sensor can be evaluated by plugging \cref{eq:lidar_rad} into \cref{eq: vol_render_nerf}: 
\begin{equation}
      P
      =\!\sum_{j=1}^N \int_{\zeta_j}^{\zeta_{j+1}}\!\!C \frac{T^2_{\zeta} \cdot \density_\zeta \reflectance_\zeta}{\zeta^2} \cos(\theta_j) \; d\zeta
      =\!\sum_{j=1}^N w_j \reflectance_{\zeta_j}',
\end{equation}
where the weights $w_j$ are now evaluated as (\cf \cref{eq:weights_nerf}):
\begin{equation}
   w_j = 2 \opacity_{\zeta_j} \cdot\prod_{k=1}^{j-1}(1 - 2 \opacity_{\zeta_k})\;.
\label{eq:lidar_weights}
\end{equation}

\subsection{Assembling the beam from multiple rays}
\label{sec:render_lidar}
Next, we apply the adapted volume rendering formulation to multiple rays within a single LiDAR beam.

\paragraph{First range estimation.}
We adopt a two-stage approach to extract range values from the neural field,\footnote{Note the similarity to the detector in the instrument that first finds the peak of the waveform, then corrects for pulse shape.} as shown in \cref{fig:overview} (a).
To estimate the range for an ideal ray $\ray$, we uniformly sample $N^c$ points and query their density values, then compute the weights $\{w^c_j\}_{j=1}^{N^c}$ using \cref{eq:lidar_weights}. A coarse peak estimate $\zeta_p$ is obtained by finding the point with the highest weight along the ray: $p = \argmax_j \{w^c_j\}_{j=1}^{N^c} $. Next, we uniformly sample $N^f$ points from the local interval $\zeta_j \in [\zeta_p - \epsilon, \zeta_p + \epsilon]$. The weights $w_j^f$ at these points are recomputed and normalized to then obtain the final, refined range estimate $\zeta_f$ as:~$\zeta_f = \sum_{j=1}^{N^f} w^f_j \cdot \zeta_j\;.$

\vspace{-3mm}
\paragraph{Second range estimation.}
As discussed in~\cref{sec:lidar_model} a single LiDAR beam might have multiple returns if enough energy was reflected from surfaces further away than the first return. To capture this behavior in our scene representation, we employ \textit{truncated} volume rendering to estimate the radiant power beyond the first return (see \cref{fig:overview} (b)).

Specifically, for each beam, we first predict a two-return mask $p_s$, by classifying its features $\rayfeat = (\geofeatbar,\dirfeat,\rangefeat)$, where $\geofeatbar$ is the volume-rendered geometric feature, and $\rangefeat$ describes the standard deviation and maximum discrepancy of range estimates at the first return. Intuitively, $\geofeatbar$ describes the local geometry (\eg an edge), $\dirfeat$ encodes the relation of the beam to the geometry, and $\rangefeat$ characterizes the beam's prior interaction with the scene.

For beams that have two returns, we then perform \textit{truncated} volume rendering as follows. We first add a buffer $\xi$\footnote{The buffer $\xi$ is sensor specific and describes the minimum spacing between two distinct returns.} to the estimated range $\zeta_1$ of the first return. We then reset the transmittance $T_{\zeta_1 + \xi}$ to 1 by zeroing out the densities up to $(\zeta_1 + \xi)$ and recalculate the weights to ensure that they adhere to \cref{eq:lidar_weights}. Finally, we repeat the range estimation described above to estimate the range of the second return $\zeta_2$. Note that for beams with two returns, the estimated range $\zeta_1$ denotes the minimum range of all rays within the beam diameter, i.e. we perform volume rendering on all rays of a beam and pick the closest one as the first return. This is different from the beams with a single return where we directly use the central ray to estimate $\zeta_1$.

\vspace{-3mm}
\paragraph{Reflectance estimation.}
At every detected surface point we can also retrieve reflectance from the neural field, using the relation $\reflectance = \sum_{j=1}^{N^f} w_j^f \cdot \reflectance_j$.

\vspace{-3mm}
\paragraph{Ray drop probability.}
In real LiDAR sensors, some emitted beams return no range measurement at all. This happens when the observed return signal has either too low amplitude or no clear peak (\cref{fig:overview} (c)). However, this effect is hard to model in a fully physics-based way,\footnote{Beyond simple thresholding, which our beam model would support.} because it depends on (usually undisclosed) details of the detection logic. We empirically observe that the ray drop probability can be learned from LiDAR measurements. To this end, we augment the neural scene representation with a dedicated variable for the local probability of \textit{not} back-scattering radiant power $\pdrop(\zeta) \in \{0, 1\}$\footnote{Please refer to the supplementary for discussions on this design choice.}. 
Volume rendering integrates that quantity into a ray drop probability: $\pdrop(\ray) = \sum_{j=1}^{N_c} w_j^c\cdot p_d(\zeta_j)$.

\subsection{Training the neural LiDAR field}
\label{sec:opt}
Given a set of posed LiDAR scans, we optimise our neural field model by minimising the loss
\begin{equation}
    \mathcal{L} = \mathcal{L}_{\text{range}} + \lambda_e \mathcal{L}_{e} + \lambda_d \mathcal{L}_{d} + \lambda_s\mathcal{L}_s\;,
\label{eq:loss}
\end{equation}
consisting of a reconstruction loss $\mathcal{L}_{\text{range}}$ for range estimation, reflectance loss $\mathcal{L}_{e}$, and classification losses $\mathcal{L}_{d}$ for ray drops and $\mathcal{L}_s$ for two returns. 

\vspace{-3mm}
\paragraph{Range reconstruction.}
We add two separate losses for the coarse range $\zeta_p$ and the refined range $\zeta_f$, $\mathcal{L}_{\text{range}} = \mathcal{L}_{\text{range}}^c + \mathcal{L}_{\text{range}}^f$. For coarse range, we impose a Gaussian distribution~\cite{rematas2021urban} around the ground truth $\hat{\zeta}$, 
\begin{equation}
    \mathcal{L}_{\text{range}}^c = \frac{1}{|\mathcal{R}|} \sum_{\ray \in \mathcal{R}} \parens*{1-\!\sum_{w_j \in \mathcal{X}_c^n}\!w_j \hat{w}_j+\!\sum_{w_k \in \mathcal{X}_c^e}\!w_k^2}\;,
\end{equation}
where $\mathcal{R}$ is the set of LiDAR rays, $\mathcal{X}_c^n$ and $\mathcal{X}_c^e$ denote points sampled within and outside the interval $[\hat{\zeta}-\epsilon,\hat{\zeta}+\epsilon]$. The ground truth weight $\hat{w}_j$ is calculated by integrating the Gaussian distribution.
The range refinement loss is defined as: $\mathcal{L}_{\text{range}}^f = \frac{1}{|\mathcal{R}|} \sum_{\ray \in \mathcal{R}} |\hat{\zeta} - \zeta_f|.$

\vspace{-3mm}
\paragraph{Reflectance reconstruction}
is optimized by minimizing an L2 loss \wrt the ground truth intensity $\hat{\intensity}$:~$\mathcal{L}_e = \frac{1}{|\mathcal{R}|} \sum_{\ray \in \mathcal{R}} (\hat{\intensity} - \intensity)^2\;.$  

\vspace{-3mm}
\paragraph{Ray drop and dual return masks} 
are trained as classification tasks, by minimizing the combination of a binary cross entropy loss $\mathcal{L}_{bce}$ and a Lovasz loss $\mathcal{L}_{ls}$~\cite{berman2018lovasz}:
\begin{equation}
     \mathcal{L}_* = \frac{1}{|\mathcal{R}|} \sum_{\ray \in \mathcal{R}} \left(\mathcal{L}_{bce}(p_*, \hat{p_*}) + \mathcal{L}_{ls}(p_*, \hat{p_*}) \right)\;.
\end{equation}
\begin{figure*}[t]
    \centering
        \includegraphics[width=1.0\textwidth]{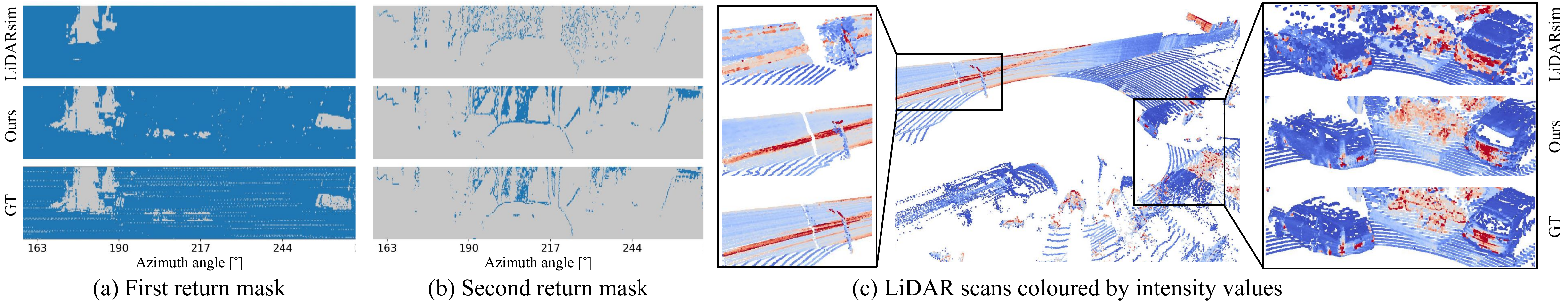}
        \caption{Qualitative results of LiDAR novel view synthesis on \textit{Waymo Interp.} dataset. On the left, we color-code rays {\setlength{\fboxsep}{0pt}\colorbox{hit}{with}} and {\setlength{\fboxsep}{0pt}\colorbox{ourgray}{without}} return. On the right side, LiDAR intensity values are color-coded as :0~\coolwarm~0.25.}
    \label{fig:two_return}
\end{figure*}
\begin{table*}[t]
    \setlength{\tabcolsep}{4pt}
    \renewcommand{\arraystretch}{1.2}
	\centering
	\resizebox{1.0\textwidth}{!}{
    \begin{tabular}{ll|ccc|ccccc|cc|ccc}
    \toprule
    & & \multicolumn{3}{c|}{First range} & \multicolumn{5}{c|}{Second range} & \multicolumn{2}{c|}{Intensity} & \multicolumn{3}{c}{Ray drop}  \\
     \multicolumn{2}{c|}{Method}    & Recall@50$\uparrow$  & MAE $\downarrow$ & MedAE $\downarrow$ & Seg. recall $\uparrow$ & Seg. precision $\uparrow$ & Recall@50 $\uparrow$ & MAE $\downarrow$ & MedAE $\downarrow$ & MAE\textsuperscript{1st} $\downarrow$ & MAE\textsuperscript{2nd} $\downarrow$ & Recall $\uparrow$ & Precision $\uparrow$ &  IoU $\uparrow$ \\
    \midrule
    \multicolumn{2}{c|}{LiDARsim~\cite{manivasagam2020lidarsim}} & 74.1 & 105.4 & 18.5 & 3.5 & 11.5 & 1.0 & 2258.0 & 1898.2 & 0.013 & 0.018 & 32.5 & \textbf{85.5} & 30.5 \\
    \arrayrulecolor{lightgray}\cline{1-15}\arrayrulecolor{black}
    \multirow{3}{*}{Ours} & Central ray & 92.8 & 32.8 & 5.6 & 79.8 & 62.9 & 61.1 & 589.1 & 21.8 & 0.004 & 0.009 & 64.3 & 81.7 & \textbf{57.1} \\
    & Diverged beam & 92.3 & 36.1 & 5.7 & 82.1 & 55.6 & 67.4 & 505.1 & 13.4 & \textbf{0.004} & \textbf{0.008} & \textbf{65.1} & 78.0 & 56.1 \\
    & GT mask & \textbf{93.2} & \textbf{29.7} & \textbf{5.6} & \textbf{100.0} & \textbf{100.0} & \textbf{79.8} & \textbf{116.0} & \textbf{8.1} & 0.004 & 0.011 & 65.1 & 78.0 & 56.1 \\
    \bottomrule
    \end{tabular}
    }
	\caption{Comprehensive ray measurement evaluation of LiDAR novel view synthesis on \textit{Waymo Interp.} dataset.}
	\label{tab:full_eval}
\end{table*}

\begin{table}[t]
    \setlength{\tabcolsep}{4pt}
    \renewcommand{\arraystretch}{1.2}
	\centering
	\resizebox{\columnwidth}{!}{
    \begin{tabular}{l|ccc|ccc|ccc|ccc}
    \toprule
    & \multicolumn{3}{c|}{TownClean} & \multicolumn{3}{c|}{TownReal} & \multicolumn{3}{c|}{Waymo interp.} & \multicolumn{3}{c}{Waymo NVS} \\
    Method  & MAE $\downarrow$ &  MedAE $\downarrow$ & CD $\downarrow$ & MAE $\downarrow$ &  MedAE $\downarrow$  & CD $\downarrow$ & MAE $\downarrow$ &  MedAE $\downarrow$ & CD $\downarrow$ & MAE $\downarrow$ &  MedAE $\downarrow$ & CD $\downarrow$ \\
    \midrule
    i-NGP~\cite{muller2022instant} & 42.2 & 4.1 & 17.4 & 49.8 & 4.8 & 19.9 & \textbf{26.4} & 5.5 & \textbf{11.6} & \textbf{30.4} & 7.3 & \underline{15.3} \\
    DS-NeRF~\cite{deng2021depth} & \underline{41.7} & 3.9 & \underline{16.6} & \underline{48.9} & 4.4 & \underline{18.8} & \underline{28.2} & 6.3 & 14.5 & 30.4 & \underline{7.2} & 16.8 \\
    URF~\cite{rematas2021urban} & 43.3 & 4.2 & 16.8 & 52.1 & 5.1 & 20.7 & 28.2 & \underline{5.4} & 12.9 & 43.1 & 10.0 & 21.2 \\
    LiDARsim~\cite{manivasagam2020lidarsim} & 159.6 & \textbf{0.8} & 23.5 & 162.8 & \underline{3.8} & 27.4 & 116.3 & 15.2 & 27.6 & 160.2 & 16.2 & 34.7 \\
    Ours & \textbf{32.0} & \underline{2.3} & \textbf{9.0} & \textbf{39.2} & \textbf{3.0} & \textbf{11.5} & 30.8 & \textbf{5.1} & \underline{12.1} & \underline{32.6} & \textbf{5.5} & \textbf{13.2} \\
    \bottomrule
    \end{tabular}
 }
	\caption{Results of LiDAR novel view synthesis for the first range.}
	\label{tab:main}
\end{table}

\begin{table}[t]
    \setlength{\tabcolsep}{4pt}
    \renewcommand{\arraystretch}{1.2}
	\centering
	\resizebox{1.0\columnwidth}{!}{
    \begin{tabular}{l|ccc|ccc}
    \toprule
    &  \multicolumn{3}{c|}{TownClean} & \multicolumn{3}{c}{Waymo Interp.} \\
    Method & MAE $\downarrow$ &  MedAE $\downarrow$ & CD $\downarrow$ & MAE $\downarrow$ & MedAE $\downarrow$ & CD $\downarrow$ \\
    \midrule
    i-NGP~\cite{muller2022instant} & 41.0 (\textcolor{green}{-1.2})& 4.1 (\textcolor{red}{0.0})& 17.6 (\textcolor{red}{0.2})& 25.3 (\textcolor{green}{-1.1})& 4.5 (\textcolor{green}{-1.0})& 10.5 (\textcolor{green}{-1.1})\\
    DS-NeRF~\cite{deng2021depth} & 37.4 (\textcolor{green}{-4.2})& 3.0 (\textcolor{green}{-0.9})& 14.4 (\textcolor{green}{-2.2})& 27.4 (\textcolor{green}{-0.8})& 5.4 (\textcolor{green}{-1.0})& 13.6 (\textcolor{green}{-0.9})\\
    URF~\cite{rematas2021urban} & 46.4 (\textcolor{red}{3.0})& 4.5 (\textcolor{red}{0.3})& 18.4 (\textcolor{red}{1.6})& 28.3 (\textcolor{red}{0.1})& 5.3 (\textcolor{green}{-0.1})& 13.1 (\textcolor{red}{0.2})\\
    Ours& 32.0 (\textcolor{green}{-2.1})& 2.3 (\textcolor{green}{-2.5})& 9.0 (\textcolor{green}{-3.9})& 30.8 (\textcolor{green}{-2.1})& 5.1 (\textcolor{green}{-2.0})& 12.1 (\textcolor{green}{-2.3})\\
    \bottomrule
    \end{tabular}
    }
	\caption{Ablation study of volume rendering for active sensing.}
	\label{tab:ablate_vol_render}
\end{table}

\begin{figure}[t]
    \centering
        \includegraphics[width=0.95\linewidth]{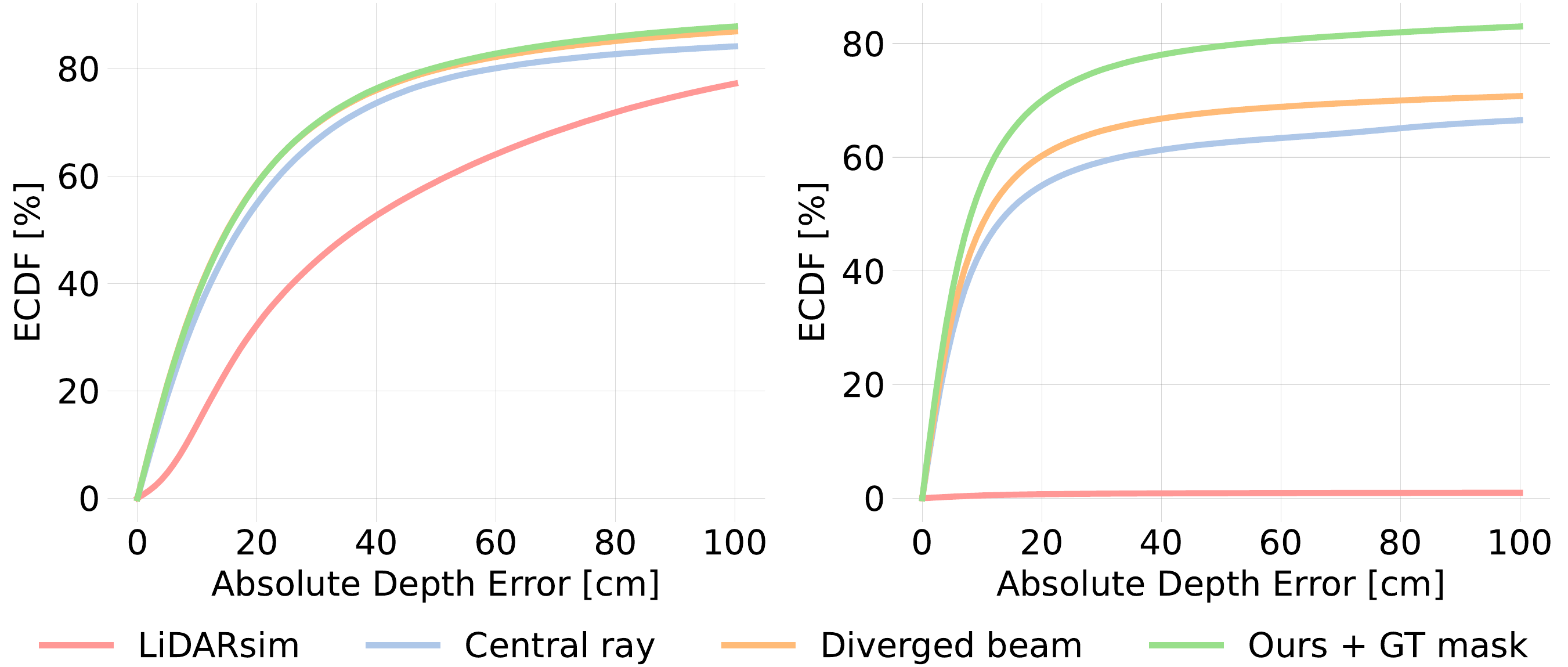}
        \caption{Beam divergence modeling improves range accuracy of rays with dual returns. This is evident in the improved error distribution of the first (left) and second return range (right).}
    \label{fig:ecdf}
\end{figure}

\begin{figure*}[t]
\centering
\includegraphics[width=1.0\textwidth]{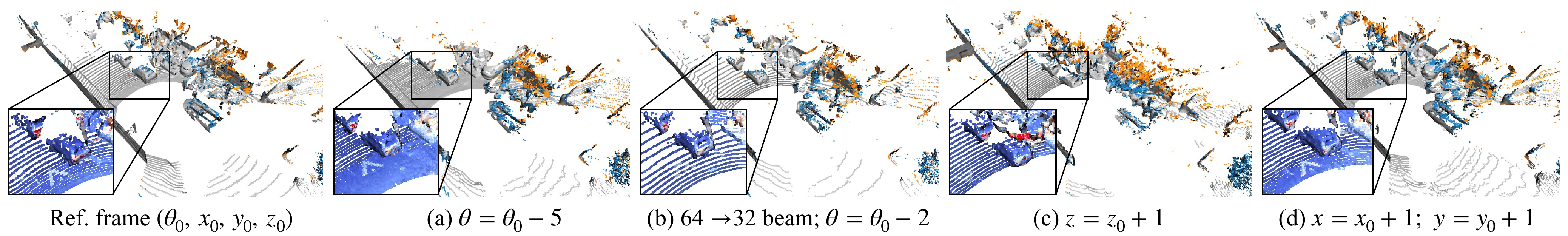}
\caption{LiDAR novel view synthesis by changing the sensor elevation angle $\theta [^{\circ}]$, pose $(x, y, z) [m]$ and number of beams. Zoom-in points are color-coded by intensity values.}
\label{fig:lidar_nvs}
\end{figure*}
\begin{figure*}[t]
\centering
\includegraphics[width=1.0\textwidth]{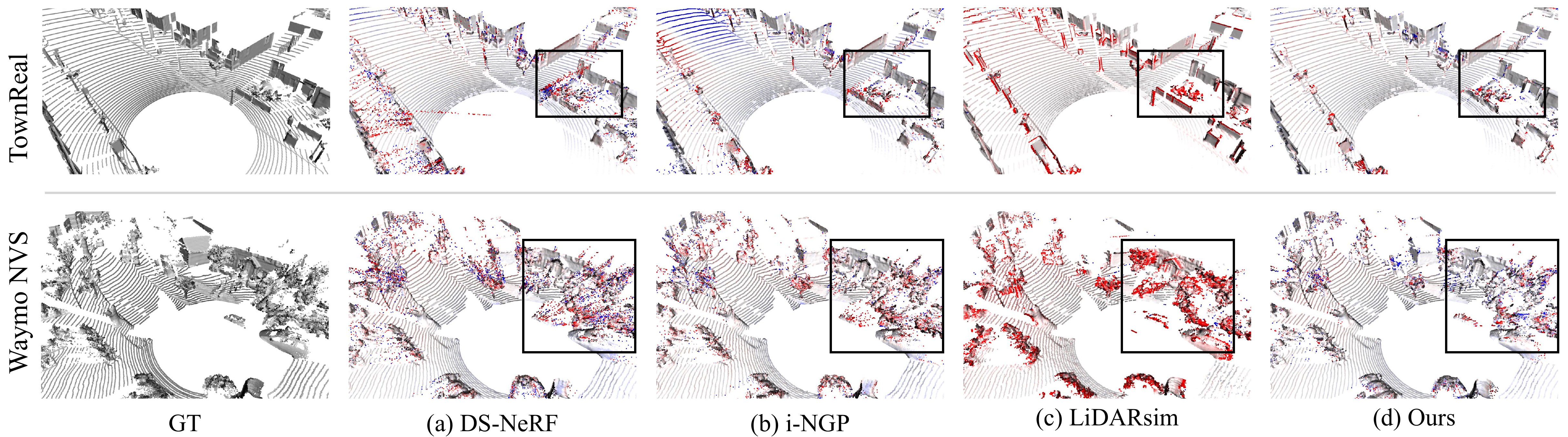}
\caption{Qualitative comparison of first range estimation. Regions with gross errors (-100 \bwr~100 cm) are highlighted.}
\label{fig:full_qualitative}
\end{figure*}

\section{Experiments}
\label{sec:results}
We start by describing our LiDAR simulator,  datasets, evaluation metrics, and baselines in \cref{sec:dataset}. In \cref{sec:lidar_nvs_eval}, we evaluate NFL directly on the LiDAR novel view synthesis task. Finally, in \cref{sec:perceptual_quality} we evaluate the suitability of our synthesized LiDAR data for two low-level tasks, point cloud registration and semantic segmentation.

\subsection{Datasets and Evaluation setting}
\label{sec:dataset}
\paragraph{LiDAR simulator -- TownReal dataset.} 
To enable quantitative evaluation in a controlled environment, we build a LiDAR simulator that allows us to virtually scan synthetic 3D assets represented either as triangular meshes or surfels. Specifically, we follow the LiDAR model described in \cref{sec:lidar_model} and allow control over the angular resolution, beam divergence, and pulse shape of the LiDAR sensor.  %

We use this simulator in combination with a 3D asset of a town~\cite{turbosquid} to synthesize the \emph{Town} dataset. We generate four scenes by splitting the 3D asset into four non-overlapping areas. Training and test scans %
are created from different trajectories.  We use two different configurations of the LiDAR sensor: \textit{(\romannumeral 1)} \textit{TownClean}, in which LiDAR scans are simulated using an idealized, non-divergent ray; and \textit{(\romannumeral 2)} \textit{TownReal}, with a diverged beam profile approximated via 37 subrays. See the supplementary material for further details.

\vspace{-3mm}
 \paragraph{Waymo Open dataset.} For evaluation on real-world data we use Waymo open dataset~\cite{sun2020scalability} which was captured by a 64-beam LiDAR sensor at 10 Hz. Here, we select four static scenes (see sequence IDs in supplementary material) and extract a five-second clip from each, resulting in 50 scans per scene. We hold out every $5$-th frame as a test view and use the remaining 40 scans for training (\textit{Waymo Interp.})
 
To evaluate the methods in a more challenging setting we propose a novel evaluation protocol based on a closed-loop simulation (\emph{Waymo NVS}). The protocol involves training and testing on all scans of a scene by first optimizing on the input views to synthesize novel views from a changed trajectory (shift the sensor by [1.5, 1.5, 0.5] meters\footnote{Please refer to the supplementary for ablations on different sensor shift configurations.}). The novel view are then used to re-optimize the method, synthesize scans in the original view and compare to the original scans to gauge performance. This formulation allows us to control task difficulty and could also be applied to evaluate camera-based novel view synthesis methods.

\vspace{-3mm}
\paragraph{Evaluation metrics.}
To evaluate range accuracy, we report four metrics: mean and median absolute errors (\textit{MAE} [cm], \textit{MedAE} [cm]),  two-way Chamfer distance (\textit{CD} [cm]), and \textit{recall@50}, which denotes the percentage of rays with range errors below 50 cm. 
We additionally measure the two return segmentation recall (\textit{Seg. recall}) and precision (\textit{Seg. precision}).
Intensity is evaluated using mean absolute error (\textit{MAE}).
For ray drop segmentation, we report recall and precision [$\%$], and intersection-over-union (\textit{IoU} [$\%$]). For point cloud registration, we report rotation error (\textit{RE} [\degree]) and translation error (\textit{TE} [cm]).

\vspace{-3mm}
\paragraph{Baselines.}
We compare NFL to four baselines. Closest to our problem setup is LiDARsim~\cite{manivasagam2020lidarsim} which was designed for LiDAR synthesis based on surface reconstruction and ray-surfel casting. We re-implement LiDARsim and augment it with a diverged beam profile to enable synthesis of the second returns. Additionally, we adapt three NeRF-like methods that were originally proposed for image synthesis i-NGP~\cite{muller2022instant}, DS-NeRF~\cite{deng2021depth}, and URF~\cite{rematas2021urban} by modifying their volumetric rendering to improve their range predictions. 
Additional details are available in the supplementary.

\subsection{Evaluation of LiDAR novel view synthesis}
\label{sec:lidar_nvs_eval}

\paragraph{Ray measurement.}

Using the \emph{Waymo Interp.} dataset we conduct a comprehensive analysis of all the ray measurements and present the results in \cref{tab:full_eval} and \cref{fig:two_return}. Only NFL and LiDARsim~\cite{manivasagam2020lidarsim} are used for this experiment, as other baselines can only support a single return. 
LiDARsim's surface representation is explicit and not optimized for novel view synthesis nor accounts for view-dependent effects, which results in inferior range prediction and difficulties in retrieving secondary returns. In contrast, NFL directly optimizes the neural field for view synthesis while accounting for LiDAR acquisition process characteristics, resulting in significantly reduced range errors and superior performance in intensity and ray drop probability estimation. Notably, equipping our model with a \textit{diverged beam} representation improves range estimation for both first and second returns for rays with dual returns(\cf ~\cref{fig:ecdf}). However, diverged beam does slightly degrade the overall first-range accuracy likely due to imprecise two-return mask estimation. This hypothesis is supported by results using the ground truth two-return mask (\textit{GT mask}). In~\cref{fig:lidar_nvs} we show more qualitative results of novel view synthesis by NFL.

\vspace{-3mm}
\paragraph{First range.}

The results of estimating the range of the first return on all datasets are presented in~\cref{tab:main} and \cref{fig:full_qualitative}. As demonstrated by the results on \textit{TownClean}, \textit{TownReal}, and \textit{Waymo NVS}, the proposed volume rendering formulation of NFL effectively regularizes the density field resulting in superior performance in challenging cases. Even in the easier setting (resembling overfitting) on \textit{Waymo Interp.} dataset, our method achieves competitive performance. In contrast, NeRF-like formulations (i-NGP~\cite{muller2022instant}, DS-NeRF~\cite{deng2021depth}, and URF~\cite{rematas2021urban}) perform poorly when evaluated on real novel views due to their inability to account for the active sensing principle. LiDARsim achieves promising results on datasets with simple geometry and clean LiDAR measurements, as evidenced by low \textit{MedAE} scores on \textit{TownClean} and \textit{TownReal}. However, its explicit representation struggles with complex geometry in noisy real-world scenes, \eg, the vegetation regions in the \textit{Waymo} dataset, resulting in high \textit{MedAE} scores.

\vspace{-3mm}
\paragraph{Ablation study of volume rendering for active sensing.}
To evaluate the effectiveness of our volume rendering formulation for active sensors, we replace the volume rendering~\cite{mildenhall2020nerf} formulation initially developed for passive sensing in all NeRF-based baselines and report performance difference in ~\cref{tab:ablate_vol_render}. Our formulation improves range accuracy across all settings, without any hyper-parameter tuning.

\begin{table}[t]
    \setlength{\tabcolsep}{4pt}
    \renewcommand{\arraystretch}{1.2}
	\centering
	\resizebox{1.0\columnwidth}{!}{
    \begin{tabular}{l|ccc|ccc|ccc}
    \toprule
    & \multicolumn{3}{c|}{TownClean} & \multicolumn{3}{c|}{TownReal} & \multicolumn{3}{c}{Waymo NVS} \\
    Method & Rec@5 $\uparrow$ & RE $\downarrow$ & TE $\downarrow$ &  Rec@5 $\uparrow$ & RE $\downarrow$ & TE $\downarrow$ & Rec@2 $\uparrow$ & RE $\downarrow$ &  TE $\downarrow$ \\
    \midrule
    i-NGP~\cite{muller2022instant} & 70.3 & 0.1 & 4.2 & 76.0 & 0.1 & 4.2 & 60.2 & 0.1 & 1.9\\
    DS-NeRF~\cite{deng2021depth} & 58.3 & 0.2 & 5.1 & 56.2 & 0.2 & 5.1 & 42.3 & 0.1 & 2.4\\
    URF~\cite{rematas2021urban} & 61.5 & 0.2 & 5.0 & 59.9 & 0.1 & 4.7 & 32.1 & 0.1 & 2.7\\
    LiDARsim~\cite{manivasagam2020lidarsim} & \textbf{82.8} & 0.1 & \textbf{3.4} & \underline{79.2} & 0.1 & \textbf{3.4} & \underline{62.8} & 0.1 & \underline{1.8}\\
    Ours & \underline{80.2} & 0.1 & \underline{3.7} & \textbf{85.9} & 0.1 & \textbf{3.4} & \textbf{71.9} & 0.1 & \textbf{1.7}\\
    
    \bottomrule
    \end{tabular}
    }
	\caption{Point cloud registration results on three datasets.}
	\label{tab:registration}
\end{table}
\begin{table}[t]
\setlength{\tabcolsep}{4pt}
\renewcommand{\arraystretch}{1.2}
\centering
\resizebox{1.0\columnwidth}{!}{
\begin{tabular}{l|ccc|ccc}
\toprule
& \multicolumn{3}{c|}{Vehicle} & \multicolumn{3}{c}{Background} \\
Method & Recall $\uparrow$ & Precision $\uparrow$ & IoU $\uparrow$ & Recall $\uparrow$ & Precision $\uparrow$ & IoU $\uparrow$ \\
\midrule
i-NGP~\cite{muller2022instant} & \underline{93.2} & 85.9 & \underline{80.9} & 98.3 & \underline{99.2} & \underline{97.6}\\
DS-NeRF~\cite{deng2021depth} & 90.7 & \textbf{87.1} & 80.2 & \textbf{98.5} & 98.9 & 97.4\\
URF~\cite{rematas2021urban} & 87.8 & 81.7 & 73.7 & 98.0 & 98.4 & 96.5\\
Lidarsim~\cite{manivasagam2020lidarsim} & 90.5 & 70.5 & 65.9 & 94.9 & 99.0 & 94.0\\
Ours & \textbf{95.9} & \underline{87.0} & \textbf{83.9} & \underline{98.3} & \textbf{99.5} & \textbf{97.8}\\
\bottomrule
\end{tabular}
}
\caption{Semantic segmentation results on \textit{Waymo NVS} dataset.}
\label{tab:sem_seg}
\end{table}
\begin{figure}[t]
    \centering
        \includegraphics[width=1.0\linewidth]{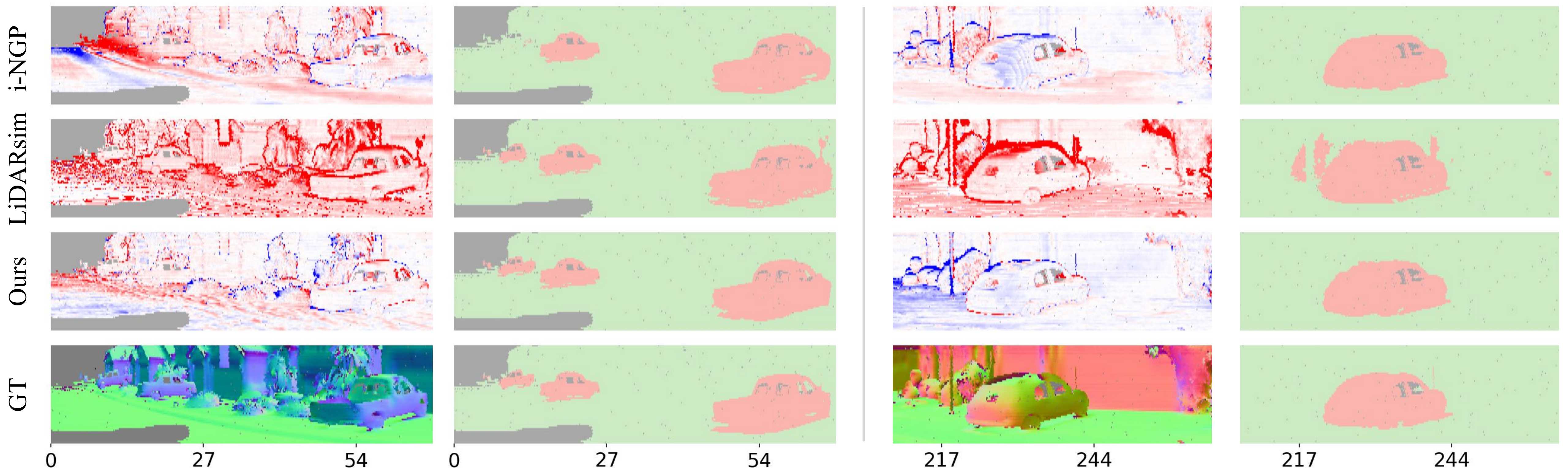}
        \caption{Semantic segmentation results on synthesised \textit{Waymo NVS} dataset. Geometry in-accuracy (-100 \bwr~100 cm) leads to erroneous semantic segmentation ({\setlength{\fboxsep}{0pt}\colorbox{ourgray}{dropped rays}}, {\setlength{\fboxsep}{0pt}\colorbox{sem0}{vehicle}}, {\setlength{\fboxsep}{0pt}\colorbox{sem1}{pedestrian}}, {\setlength{\fboxsep}{0pt}\colorbox{sem2}{background}}).}
    \label{fig:semseg}
\end{figure}

\subsection{Downstream evaluation of novel views}
\label{sec:perceptual_quality} 
Having demonstrated NFL's improved ability to synthesize high-quality LiDAR scans through various metrics, we proceed to evaluate their perceptual quality by using them as input for two low-level perception tasks: point cloud registration~\cite{huang2021predator} and semantic segmentation~\cite{tang2020searching}.

\vspace{-3mm}
\paragraph{Point cloud registration.}
To evaluate the extent to which synthesized scans preserve local geometric features, we apply the same point cloud registration model~\cite{huang2022dynamic} pre-trained on Waymo~\cite{sun2020scalability} to both GT LiDAR scans and scans synthesized using different methods. \cref{tab:registration} shows that NFL outperforms the baseline methods on datasets with complex geometry and higher noise levels (\textit{TownReal} and \textit{Waymo NVS}) that are more susceptible to artifacts occurring as a result of the LiDAR acquisition process.

\vspace{-3mm}
\paragraph{Semantic segmentation.}
To probe the potential domain gap between real and synthetic scans we apply the same, pre-trained semantic segmentation model~\cite{tang2020searching} to both and compare the predictions. \cref{tab:sem_seg} depicts the performance for both the \textit{vehicle} and \textit{background} classes. Notably, NFL achieves the highest recall for the \textit{vehicle} class, which is strongly affected by dual returns and ray drops. Example predictions are shown in~\cref{fig:semseg}.

\section{Limitations and future work}
\label{sec:conclusions}
We have presented NFL, a neural field-based approach for synthesizing LiDAR scans from novel viewpoints. NFL combines the benefits of volume rendering with a physically based model of LiDAR acquisition process to faithfully model LiDAR characteristics including beam divergence, secondary returns, and ray dropping. Even though NFL significantly outperforms explicit reconstruct-then-simulate methods as well as other NeRF-style methods, it still has some limitations that we would like to address in future work. Firstly, with NFL, we try to seek a balance between adhering to the physical principles of LiDAR and incorporating semantic features and learning. While our formulation already shows improved performance over baselines, there is still potential for further improvements. For example, while real-world LiDAR sensors perform range detection on the integrated beam radiant, we actually found that using the density-based weights separately for each ray leads to improved performance. Prediction of the second return mask enables us to model secondary returns and to further improve the estimation of the first return. Yet, as indicated by the oracle study, improving the mask prediction could lead to further improvements. Finally, our method is based on a NeRF-style representation and therefore requires per-scene optimization. Generalization across scenes and handling dynamic environments are key challenges that we plan to address in future work.

\paragraph{Acknowledgements.}
{We sincerely thank Benjamin Naujoks, Steven Butrimas, Tomislav Medić, Yu Han, and Prof.~Dr.~Andreas Wieser for helpful discussions around LiDAR models. We are grateful for the feedback on figures from Rodrigo Caye Daudt. This appreciation extends to Zvi Greenstein for organisation support.}

\clearpage
{\small
\bibliographystyle{ieee_fullname}
\bibliography{egbib.bib}
}

\clearpage
\setcounter{section}{0}
\renewcommand\thesection{\Alph{section}}
\newcommand{\manuallabel}[2]{\def\@currentlabel{#2}\label{#1}}
\makeatother
\manuallabel{eq:loss_function}{Eq.~(15)}
\newcommand{\refpaper}[1]{{\hypersetup{linkcolor={red}}\ref{#1}}}

In this supplementary document, we first present additional information about our dataset, evaluation setting, implementation details in \cref{sec:supp_data}. We then elaborate on technical details of our methods in \cref{sec:supp_method}. Additional results of the two return mask segmentation, more quantitative and qualitative results are provided in \cref{sec:supp_results}.

\section{Datasets and implementation details}
\label{sec:supp_data}
\subsection{Dataset}
\begin{figure}[t]
\centering
\includegraphics[width=1.0\columnwidth]{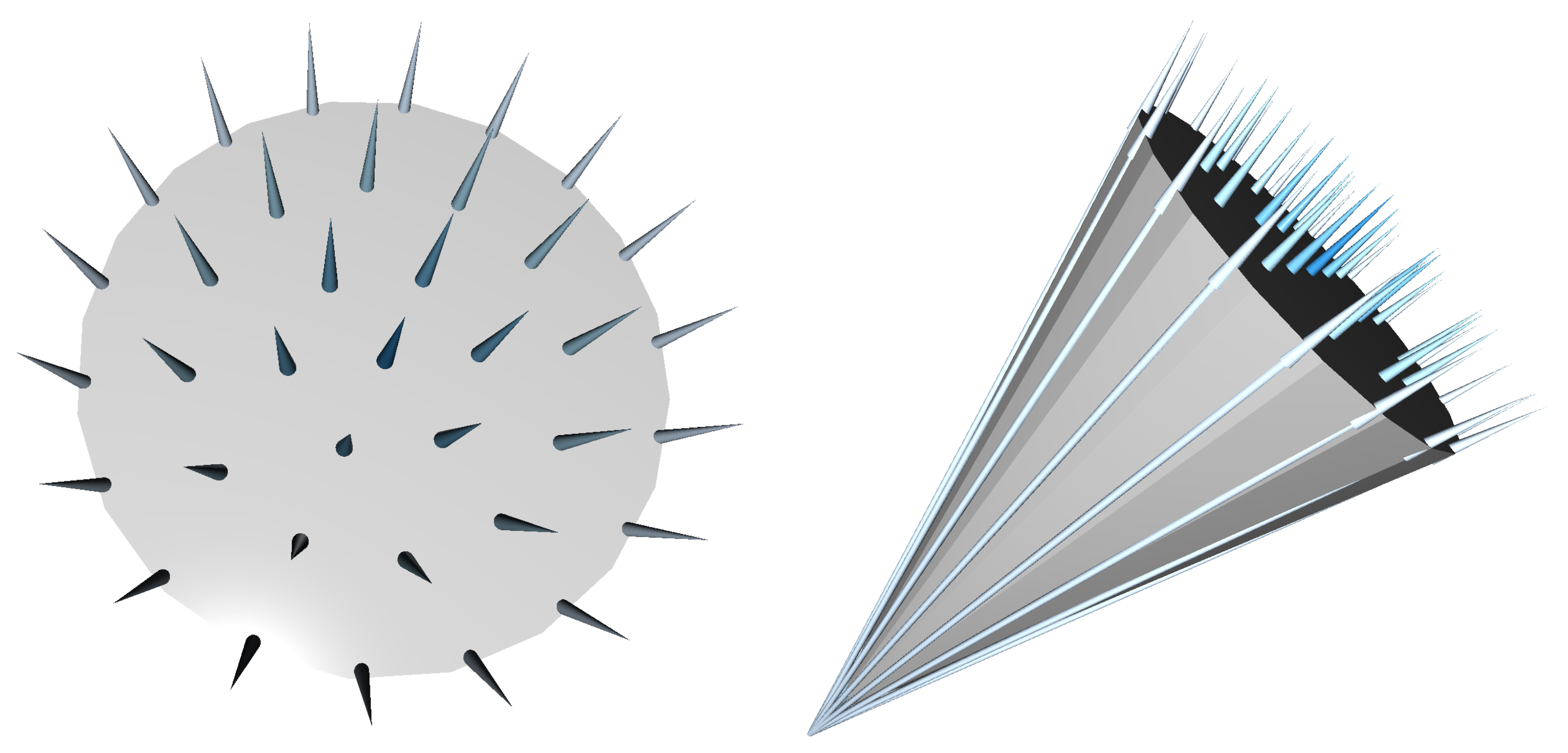}
\caption{Example diverged beam profile approximated via 37 diverged rays.}
\label{fig:diverged_ray}
\end{figure}

\begin{figure*}[t]
\centering
\includegraphics[width=0.8\textwidth]{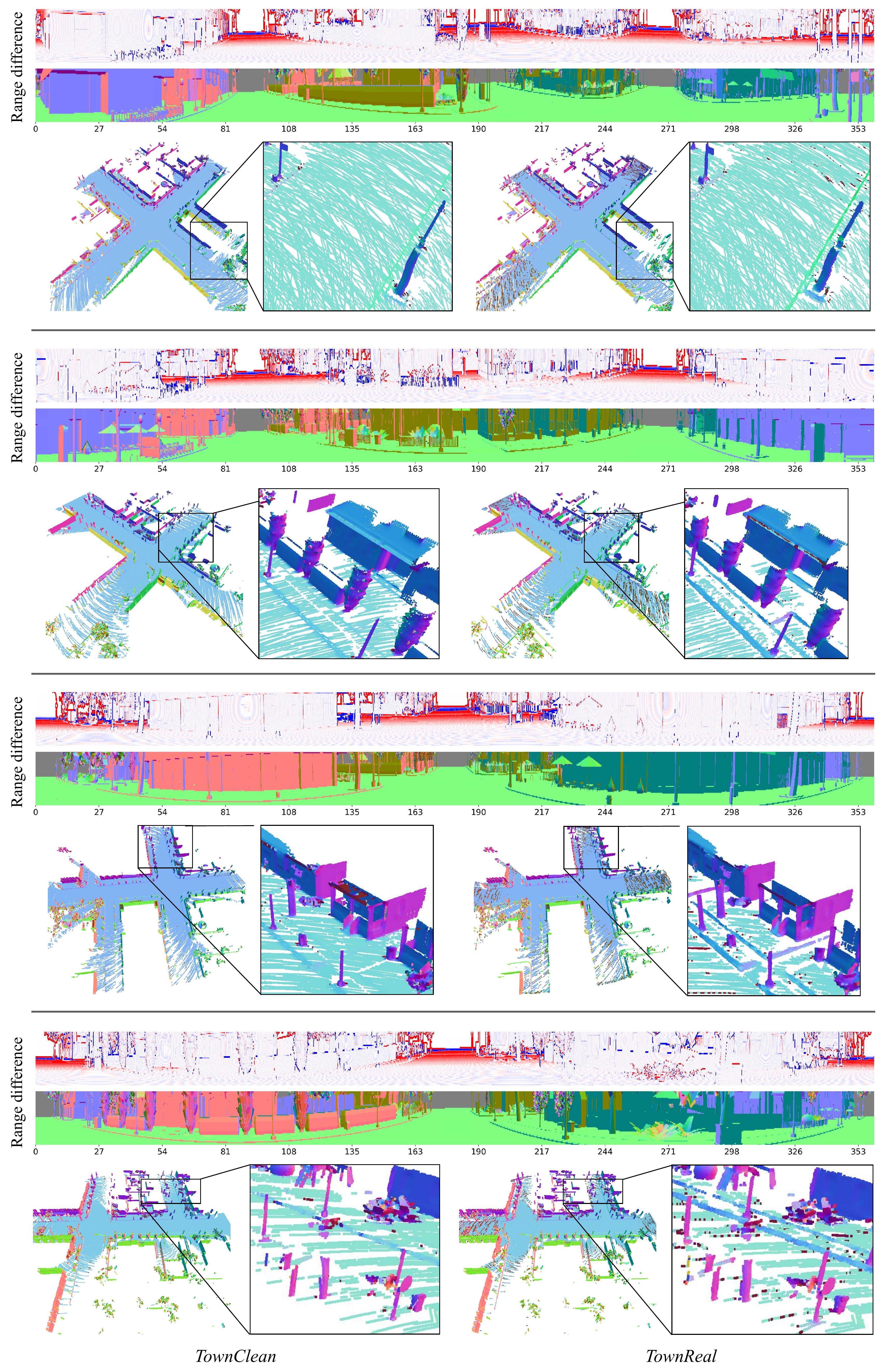}
\vspace{-3mm}
\caption{Visualisation of \textit{Town} dataset. Employing a diverged beam profile in range simulation results in an overestimation of range in the high range regime (-16 \bwr~16 cm). Such range difference is also reflected on delicate structures, as evidenced by the point cloud view.}
\label{fig:supp_town_dataset}
\vspace{-3mm}
\end{figure*}

\paragraph{Town dataset}
To simulate \textit{TownReal} dataset, we approximate a diverged beam profile using 37 subrays and the divergence angle $\gamma_0 = 2 $~mrad~\cite{glennie2012calibration}. We use the subray distribution proposed from~\cite{winiwarter2022virtual}~(\cf \cref{fig:diverged_ray}). The dataset is shown in~\cref{fig:supp_town_dataset}. 

\paragraph{Waymo dataset}
We use the following 4 scenes~(\cf \cref{fig:supp_waymo_dataset}) that are mostly static from \textit{Waymo}~\cite{sun2020scalability} dataset

\begin{table}[!h]
    \setlength{\tabcolsep}{4pt}
    \renewcommand{\arraystretch}{1.2}
	\centering
	\resizebox{0.8\columnwidth}{!}{
    \begin{tabular}{l|c}
    \toprule
    & Scene ID \\
    \midrule
    Scene 1 & 10017090168044687777\_6380\_000\_6400\_000 \\
    Scene 2 & 10096619443888687526\_2820\_000\_2840\_000 \\
    Scene 3 & 10061305430875486848\_1080\_000\_1100\_000 \\
    Scene 4 & 10275144660749673822\_5755\_561\_5775\_561 \\
    \bottomrule
    \end{tabular}
    }
\end{table}

\subsection{Evaluation setting}
\begin{figure*}[t]
\centering
\includegraphics[width=1.0\textwidth]{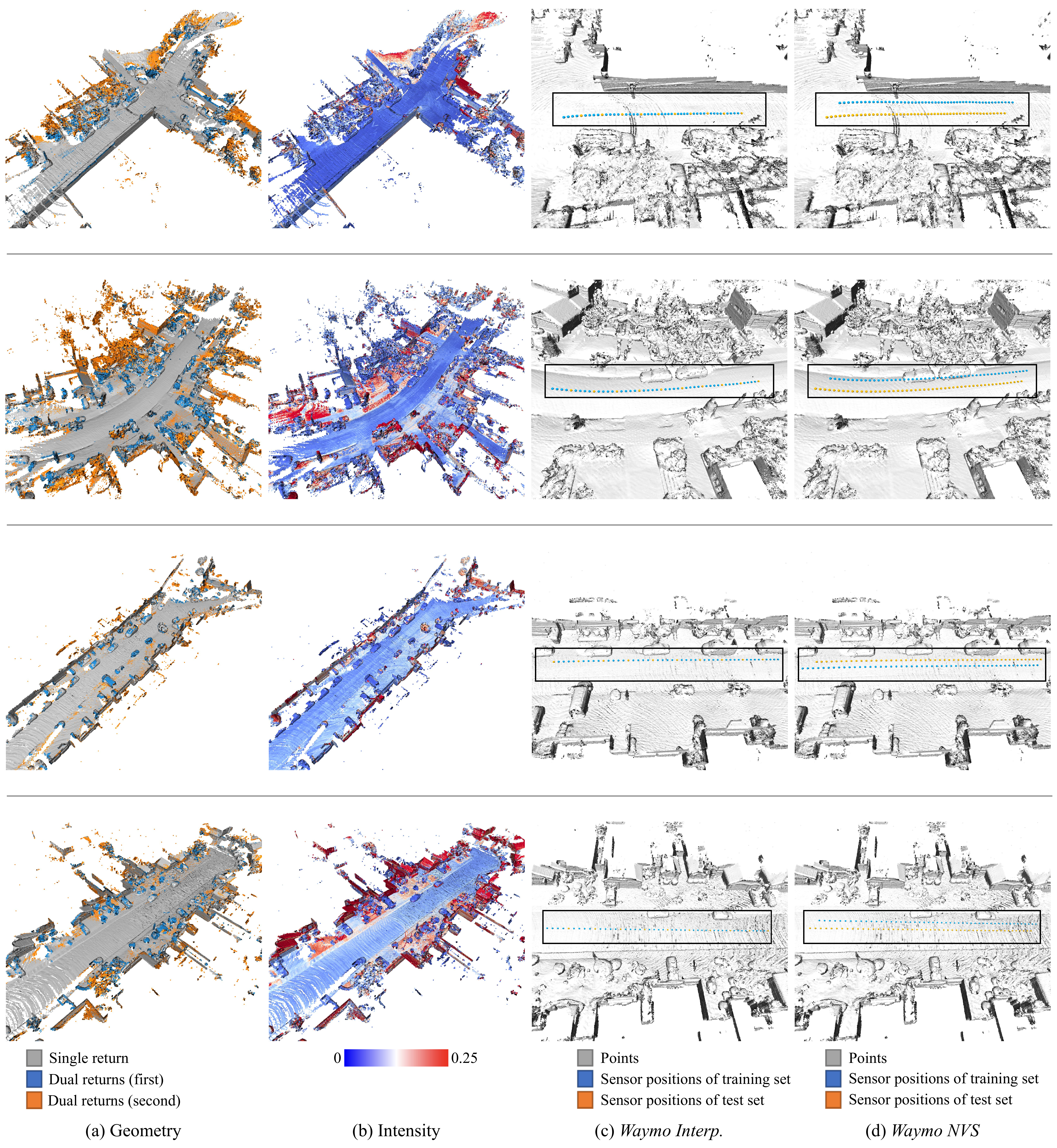}
\vspace{-3mm}
\caption{Visualisations of \textit{Waymo} dataset. We accumulate all 50 frames for each scene and show their geometry, intensity profile, and sensor positions of training and test sets on \textit{Waymo Interp.} and \textit{Waymo NVS} datasets.}
\label{fig:supp_waymo_dataset}
\vspace{-3mm}
\end{figure*}

\paragraph{\textit{Waymo NVS} setting} 
We simulate the new trajectory by shifting the sensor by [1.5, 1.5, 0.5] meters (see \cref{fig:supp_waymo_dataset}), yielding an overall displacement of ${\approx}2.18$ meters. This displacement magnitude corresponds to the requirements of various tasks, such as lane changes or adapting the sensor rig from a car to a truck. Moreover, our displacement from the trajectory is similar~\cite{Yang_2023_unisim} or even larger~\cite{Ost_2022_CVPR} than used in prior NVS works. Nevertheless, we run additional experiments by varying the displacements and report results in \cref{tab:rebuttaL_nvs}. NFL consistently outperforms baseline methods under different settings, and the improvement is more pronounced under large displacements.

\begin{table}[t]
    \captionsetup{font=scriptsize}
    \setlength{\tabcolsep}{4pt}
    \renewcommand{\arraystretch}{1.2}
	\centering
	\resizebox{1.0\columnwidth}{!}{
    \begin{tabular}{c|ccccc}
    \toprule
     & i-NGP & DS-NeRF & URF & LiDARsim & Ours \\
    \midrule
    (0.5, 0.5, 0.5)  & 7.0 / 14.4 & 7.0 / 16.0 & 9.0 / 19.6 & 16.1 / 33.1 & \textbf{5.4} / \textbf{13.0} \\
    (1.5, 1.5, 1.0)  & 8.4 / 17.6 & 7.8 / 18.5 & 11.0 / 27.5 & 16.5 / 37.9 & \textbf{5.8} / \textbf{14.3}  \\
    (2.5, 2.5, 1.5)  & 11.6 / 28.0 & 9.3 / 22.8 & 13.9 / 35.5 & 17.2 / 46.3 & \textbf{6.4} / \textbf{18.4} \\
    \bottomrule
    \end{tabular}
    }
    \caption{Varying the displacement on \textit{Waymo NVS} dataset. Numbers are reported as \textit{MedAE} / \textit{CD} [cm].}
    \label{tab:rebuttaL_nvs}
\end{table}

\paragraph{Point cloud registration task}
We utilize 49 paired consecutive frames per scene, with a relative displacement of ${\approx}1$ meter. \textit{TE} is reported in centimeters and \textit{RE} is reported in degrees.

\subsection{Implementation details}
\paragraph{Our method.}
Our model is implemented based on \emph{torch-ngp}~\cite{torch-ngp,muller2022instant} and can be trained on a single RTX 3090 GPU. During training we minimize \refpaper{eq:loss_function} using the Adam~\cite{kingma2014adam} optimiser, with an initial learning rate of 0.005 which linearly decays to 0.0005 towards the end of training. We clip the gradient magnitudes of all parameters to 1.0 to stabilize the optimisation. In the first stage, we sample $N^c = 768$ points and in the second stage $N^f=64$ points for each ray. The window size $\epsilon$ for volume rendering is set to 0.8 m, and the buffer value $\xi$ between two returns is set to 2 m. The weights in the loss function, i.e., $\lambda_e$, $\lambda_d$, and $\lambda_s$, are set to 50, 0.15, and 0.15, respectively.

\begin{figure*}[t]
\centering
\includegraphics[width=1.0\textwidth]{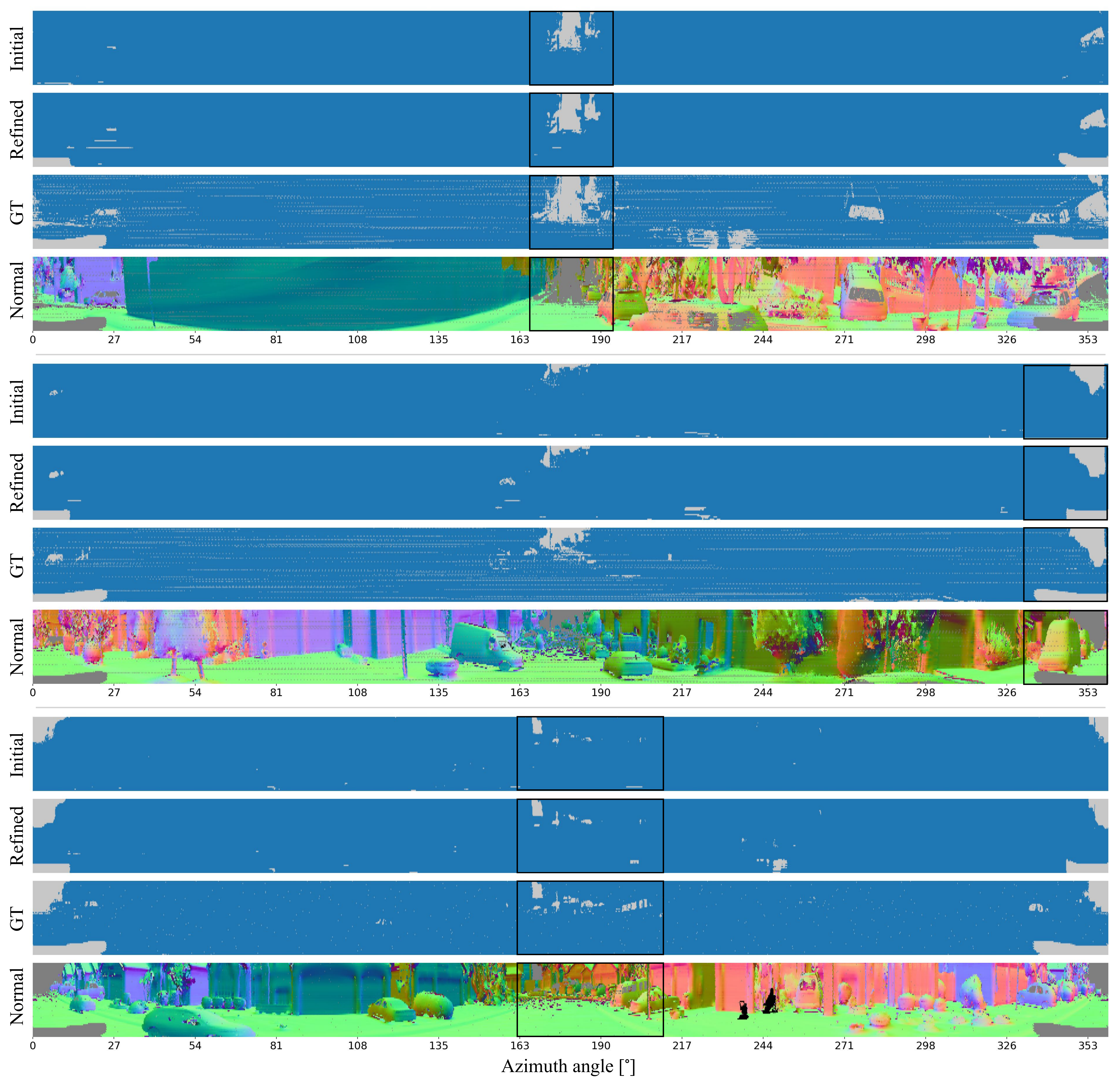}
\vspace{-3mm}
\caption{Ray drop segmentation on \textit{Waymo Interp.} dataset using LiDARsim~\cite{manivasagam2020lidarsim}. We show both the initial ray drop mask from ray-surfel query and the refined masks using learned ray-drop model.}
\label{fig:supp_lidarsim_raydrop}
\vspace{-3mm}
\end{figure*}

\paragraph{LiDARsim.}
Because the original implementation is not publicly available, we re-implemented LiDARsim~\cite{manivasagam2020lidarsim} following the paper as close as possible. Specifically, for all points in the training set, we first estimate pointwise normal vectors using all points within a 20 cm radius ball. Then, we apply voxel down-sampling~\cite{tang2022torchsparse} with a voxel size of 4 cm and reconstruct a disk surfel\footnote{We use the implementation from Point-Cloud-Utils~\cite{point-cloud-utils} library.} for each point. Here, the input point represents the disk center and it orientation is defined by the estimated normal vector. At inference time, we perform ray-surfel intersection to determine the intersection points. We empirically observed that LiDARsim's~\cite{manivasagam2020lidarsim} performance is sensitive to the selected surfel radius. Therefore, we have experimented with both a distance-dependent and fixed surfel radius and found that fixed surfel radius of 6 cm and 12 cm for \textit{Waymo} and \textit{Town} dataset, respectively lead to best range accuracy. To enable second range estimation, we augment LiDARsim with a diverged beam profile approximated using 7 rays. To obtain the second return mask, we consider a LiDAR beam to have two returns if the maximum range difference between all subrays is larger than a threshold\footnote{Sensor-specific parameter, 2 m on \textit{Waymo} dataset.}. The first return is defined as the closest ray-surfel intersection, while the second return is the nearest one that is at least two meters away. To train the ray drop module, we utilize 40k samples from the Waymo dataset~\cite{sun2020scalability}, and only apply this module after the ray-surfel intersection to refine the ray drop patterns. Please see~\cref{fig:supp_lidarsim_raydrop} for more qualitative results.

\paragraph{Other NeRF methods.}
We also use \emph{torch-ngp}~\cite{torch-ngp} codebase to implement other methods, using the same network and sampling configurations as used in ours. To estimate the range, we remove the radiance MLP and instead, apply volume rendering of the sampled $\zeta$ along the ray.  For DS-NeRF~\cite{deng2021depth} and URF~\cite{rematas2021urban}, we replace their positional encoding with a hash-grid~\cite{muller2022instant} to facilitate a fair comparison with i-NGP~\cite{muller2022instant}. Moreover, we substitute the original L2 loss with the L1 loss, as it results in better performance. Finally, we follow the original paper and augment DS-NeRF~\cite{deng2021depth} and URF~\cite{rematas2021urban} with the ray distribution loss and line-of-sight loss, respectively, to regularise the underlying geometry.

\section{Methodology and loss functions}
\label{sec:supp_method}
\paragraph{First range estimation}
If the maximum weight at the first stage $w_p^c$ is below a predefined threshold $\eta=0.1$, we assume that the network is uncertain about the reconstruction and the resulting range estimate may be inaccurate. In these cases, we only apply the coarse stage volume rendering and directly estimate the range as: $\zeta = \sum_{j=1}^{N^c} w_j^c \cdot \zeta_j$. 

\paragraph{Range reconstruction loss}
For coarse range, we impose a Gaussian distribution around the ground truth $\hat{\zeta}$ and we anneal the standard deviation $\delta$ during training, the annealing procedure is defined as:
\begin{equation}
    \delta_k = \delta_{\max} \left(\frac{\delta_{\min}}{\delta_{\max}}\right)^{k / k_{\max}}
\end{equation}
where $k$ denotes the iteration number, $k_{\max}$ is the maximum iteration, and $\delta_{\max}$ and $\delta_{\min}$ correspond to empirically determined bounds for the standard deviation. The annealing parameters $\delta_{\min}$ and $\delta_{\max}$ are set to 0.25/0.3 and 1.2/1.6, respectively, for the \textit{Town} and \textit{Waymo} datasets. The maximum iteration $k_{\max}$ is set to 16000/24000 for the \textit{Town} and \textit{Waymo} datasets. 
The ground truth weight $\hat{w}_j$ is computed as:
\begin{equation}
    \hat{w}_j = \int_{\zeta_j}^{\zeta_{j+1}} \frac{1}{\delta\sqrt{2\pi}}\exp\left(-\frac{(x - \hat{\zeta})^2}{2\delta^2}\right) \; dx.
\end{equation}

\section{Additional results}
\label{sec:supp_results}
\begin{figure*}[!t]
    \captionsetup{font=scriptsize}
    \centering
        \includegraphics[width=1.0\textwidth]{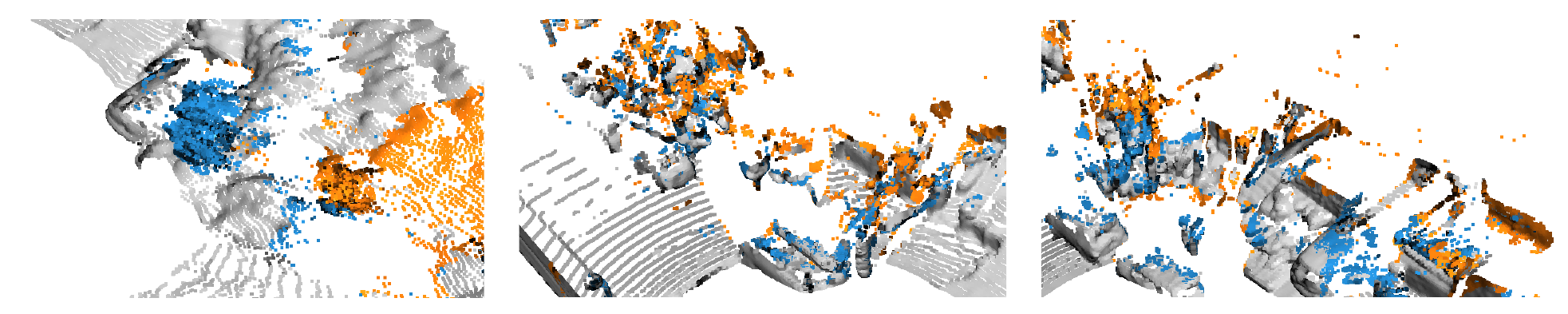}
        \caption{Rendered secondary returns are color-coded in {\setlength{\fboxsep}{0pt}\colorbox{sdpoints}{yellow}}.}
    \label{fig:rebuttal_second_return}
\end{figure*}
\paragraph{Runtime analysis}
Our \textit{central ray} version takes 4.1 ms per frame to render the single returns on an RTX 3090 GPU, while other NeRF-style methods require 2.4 ms. Only around $10\%$ of rays have second returns, resulting in low computational overhead. While our \textit{diverged beam} incurs additional costs due to querying diverged rays, it can be disabled if needed, without compromising first return performance (\cf Tab. 1). Our re-implementation of LiDARsim achieves 10 Hz runtime, but could be further improved using accelerated ray-tracing, \eg OptiX. Note that all methods already match or even (greatly) exceed the normal LiDAR measurement frequency (${\approx}10$ Hz). 

\paragraph{Ray drop modelling}
There clearly is a link between ray drops and beam divergence. However, we found that modeling it through the beam feature yields worse performance, possibly because $\rayfeat$ uses $\rangefeat$, which encodes the statistics of returns and is less meaningful for dropped rays. In future work, beam divergence could instead be incorporated through Intergrated Positional Encoding~\cite{barron2023zipnerf} to model ray drops.

\begin{figure*}[t]
\centering
\includegraphics[width=1.0\textwidth]{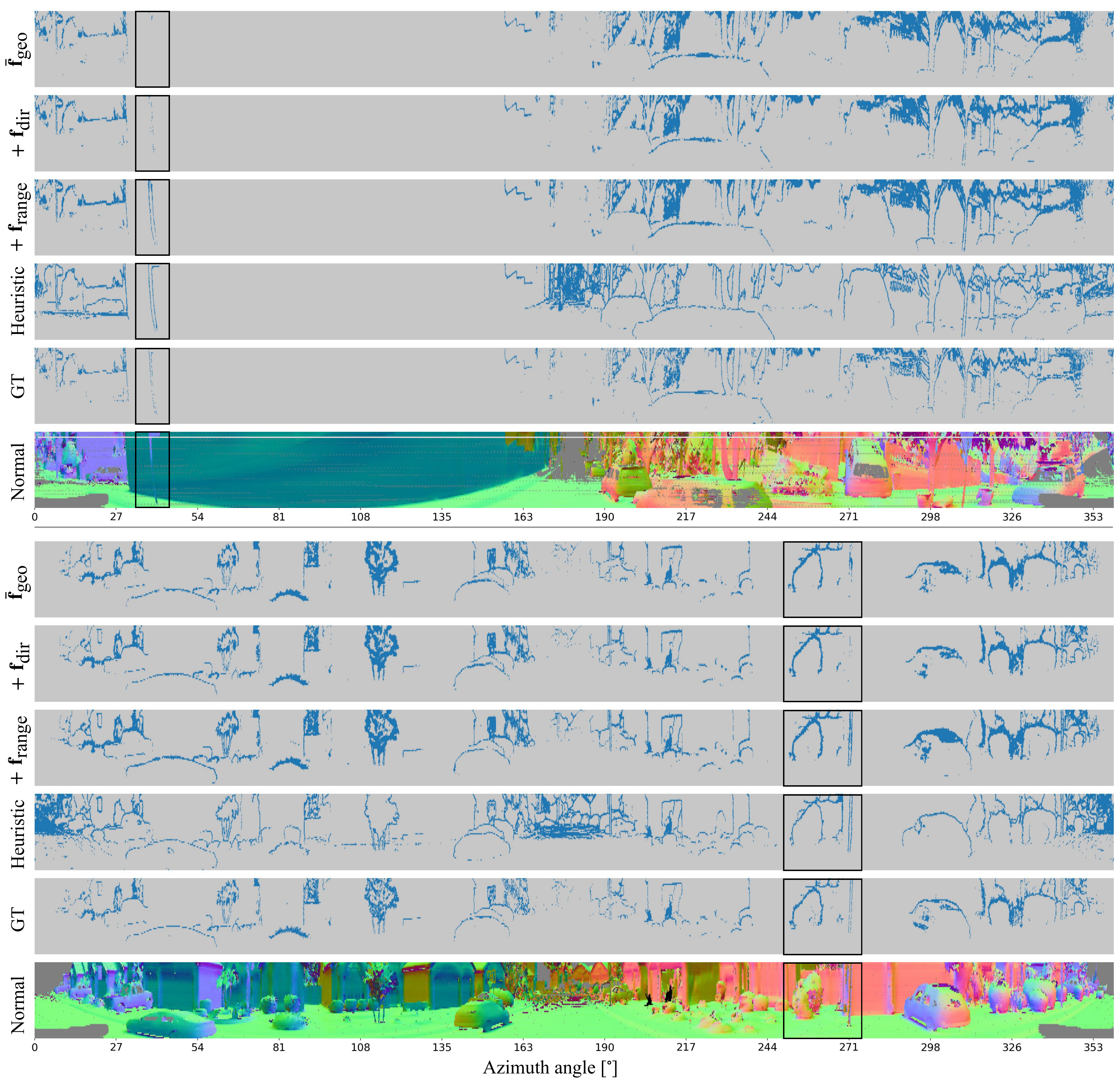}
\vspace{-3mm}
\caption{Qualitative results of two return mask segmentation.}
\label{fig:supp_ablate_two_return_mask}
\vspace{-3mm}
\end{figure*}

\paragraph{Two return mask prediction}
\begin{table}[t]
    \setlength{\tabcolsep}{4pt}
    \renewcommand{\arraystretch}{1.2}
	\centering
	\resizebox{\columnwidth}{!}{
    \begin{tabular}{ccc|ccc|ccc}
    \toprule
    \multicolumn{3}{c|}{Features} & \multicolumn{3}{c|}{Two return segmentation} & \multicolumn{3}{c}{Second range} \\
     $\overline{\textbf{f}}_\text{geo}$ & $\dirfeat$ & $\rangefeat$ & Recall $\uparrow$ & Precision $\uparrow$ & IoU $\uparrow$ & Recall@0.5 $\uparrow$ & MAE $\downarrow$ & MedAE $\downarrow$ \\
    \midrule
    \ding{51} & & & 78.0 & 61.6 & 52.8 & 60.1 & 620.1 & 26.7 \\
    \ding{51} & \ding{51} & & 79.8 & 62.9 & 54.5 & 61.1 & 589.1 & 21.8 \\
    \ding{51} & \ding{51} & \ding{51} & 82.1 & 55.6 & 49.8 & 67.4 & 505.1 & 13.4 \\
    \arrayrulecolor{lightgray}\cline{1-9}\arrayrulecolor{black}
    \multicolumn{3}{c|}{threshold depth std.} & 30.8 & 24.2 & 14.8 & 24.7 & 1532.2 & 1461.4 \\
    \bottomrule\textbf{}
    \end{tabular}
    }
	\caption{Qualitative results of two return segmentation on \textit{Waymo Interp}. dataset.}
	\label{tab:ablate_two_return_mask}
\end{table}
We conduct an ablation study to investigate different design choices for predicting the two return mask and summarize the results in~\cref{tab:ablate_two_return_mask}. We observe that concatenating the range feature $\rangefeat$ with the beam feature $\rayfeat$ improves the segmentation recall and, consequently, the second range estimation. In addition to predicting the two return mask from the beam feature, we experiment with a simple heuristic-based baseline that thresholds the depth standard deviation of sub-rays. Specifically, we considered a LiDAR beam to have two returns if the standard deviation is above 30\footnote{Empirically determined as it leads to the best Intersection-of-Union score.} cm. However, as shown in Table~\ref{tab:ablate_two_return_mask}, this approach achieves limited success and performs much worse than the learned methods. More qualitative results are presented in~\cref{fig:supp_ablate_two_return_mask}. 

\paragraph{Importance of the second return}
Multiple returns are critical for vegetation analysis in remote sensing~\cite{lim2003lidar}. NFL is the first work to model the second return by combining beam divergence and \textit{truncated} volume rendering. Unfortunately, second returns do not have semantic annotations in the Waymo dataset, which precluded a quantitative analysis. Nevertheless, qualitatively the rendered second returns are located mostly in vegetation regions, as shown in \cref{fig:rebuttal_second_return}. This correlation suggests that secondary returns could indeed be useful for detecting vegetation.

\paragraph{Semantic segmentation on \textit{Waymo Interp.} dataset}

\begin{table}[t]
\setlength{\tabcolsep}{4pt}
\renewcommand{\arraystretch}{1.2}
\centering
\resizebox{0.8\columnwidth}{!}{
\begin{tabular}{l|ccc|ccc}
\toprule
& \multicolumn{3}{c|}{Vehicle} & \multicolumn{3}{c}{Background} \\
Method & Recall $\uparrow$ & Precision $\uparrow$ & IoU $\uparrow$ & Recall $\uparrow$ & Precision $\uparrow$ & IoU $\uparrow$ \\
\midrule
i-NGP~\cite{muller2022instant} + L2 & 71.1 & \textbf{97.0} & 69.4 & \textbf{99.6} & 96.5 & 96.2\\
i-NGP~\cite{muller2022instant} & \underline{94.8} & 89.7 & \underline{85.6} & 98.7 & \underline{99.4} & \underline{98.1}\\
DS-NeRF~\cite{deng2021depth} & 91.4 & 88.9 & 82.2 & 98.7 & 99.1 & 97.8\\
URF~\cite{rematas2021urban} & 93.8 & 89.0 & 84.1 & 98.6 & 99.3 & 97.9\\
Lidarsim~\cite{manivasagam2020lidarsim} & 92.2 & 74.4 & 70.2 & 95.9 & 99.1 & 95.1\\
Ours & \textbf{95.7} & \underline{91.2} & \textbf{87.6} & \underline{98.8} & \textbf{99.5} & \textbf{98.3}\\
\bottomrule
\end{tabular}
}
\vspace{-3mm}
\caption{Semantic segmentation results on \textit{Waymo Interp.} dataset.}
\label{tab:supp_sem_seg_interp}
\end{table}
We report additional semantic segmentation results on \textit{Wamo Interp.} dataset in \cref{tab:supp_sem_seg_interp}. NFL achieves the best performance for vehicle segmentation. Please note that \textit{Waymo Interp.} is of significantly smaller size (10 test frames \vs 50 frames per scene in other datasets).

\paragraph{Quantitative results}
\begin{table}[t]
    \setlength{\tabcolsep}{4pt}
    \renewcommand{\arraystretch}{1.2}
	\centering
	\resizebox{\columnwidth}{!}{
    \begin{tabular}{l|ccc|ccc|ccc|ccc}
    \toprule
    & \multicolumn{3}{c|}{TownClean} & \multicolumn{3}{c|}{TownReal} & \multicolumn{3}{c|}{Waymo interp.} & \multicolumn{3}{c}{Waymo NVS} \\
    Method  & MAE $\downarrow$ &  MedAE $\downarrow$ & CD $\downarrow$ & MAE $\downarrow$ &  MedAE $\downarrow$  & CD $\downarrow$ & MAE $\downarrow$ &  MedAE $\downarrow$ & CD $\downarrow$ & MAE $\downarrow$ &  MedAE $\downarrow$ & CD $\downarrow$ \\
    \midrule
    i-NGP~\cite{muller2022instant} + L2 & 63.6 & 14.8 & 37.1 & 78.2 & 18.4 & 44.5 & 41.4 & 14.7 & 24.9 & 47.3 & 17.6 & 29.5 \\
    i-NGP~\cite{muller2022instant} & 42.2 & 4.1 & 17.4 & 49.8 & 4.8 & 19.9 & \textbf{26.4} & 5.5 & \textbf{11.6} & \textbf{30.4} & 7.3 & \underline{15.3} \\
    DS-NeRF~\cite{deng2021depth} & \underline{41.7} & 3.9 & \underline{16.6} & \underline{48.9} & 4.4 & \underline{18.8} & \underline{28.2} & 6.3 & 14.5 & 30.4 & \underline{7.2} & 16.8 \\
    URF~\cite{rematas2021urban} & 43.3 & 4.2 & 16.8 & 52.1 & 5.1 & 20.7 & 28.2 & \underline{5.4} & 12.9 & 43.1 & 10.0 & 21.2 \\
    LiDARsim~\cite{manivasagam2020lidarsim} & 159.6 & \textbf{0.8} & 23.5 & 162.8 & \underline{3.8} & 27.4 & 116.3 & 15.2 & 27.6 & 160.2 & 16.2 & 34.7 \\
    Ours & \textbf{32.0} & \underline{2.3} & \textbf{9.0} & \textbf{39.2} & \textbf{3.0} & \textbf{11.5} & 30.8 & \textbf{5.1} & \underline{12.1} & \underline{32.6} & \textbf{5.5} & \textbf{13.2} \\
    \bottomrule
    \end{tabular}
    }
	\caption{Results of LiDAR novel view synthesis for the first range.}
	\label{tab:supp_main}
\end{table}
\begin{table}[t]
    \setlength{\tabcolsep}{4pt}
    \renewcommand{\arraystretch}{1.2}
	\centering
	\resizebox{0.9\columnwidth}{!}{
    \begin{tabular}{l|ccc|ccc}
    \toprule
    &  \multicolumn{3}{c|}{TownClean} & \multicolumn{3}{c}{Waymo Interp.} \\
    Method & MAE $\downarrow$ &  MedAE $\downarrow$ & CD $\downarrow$ & MAE $\downarrow$ & MedAE $\downarrow$ & CD $\downarrow$ \\
    \midrule
    i-NGP~\cite{muller2022instant} + L2& 60.8 (\textcolor{green}{-2.8})& 12.6 (\textcolor{green}{-2.2})& 34.4 (\textcolor{green}{-2.7})& 40.8 (\textcolor{green}{-0.6})& 13.1 (\textcolor{green}{-1.6})& 24.0 (\textcolor{green}{-0.8})\\
    i-NGP~\cite{muller2022instant} & 41.0 (\textcolor{green}{-1.2})& 4.1 (\textcolor{red}{0.0})& 17.6 (\textcolor{red}{0.2})& 25.3 (\textcolor{green}{-1.1})& 4.5 (\textcolor{green}{-1.0})& 10.5 (\textcolor{green}{-1.1})\\
    DS-NeRF~\cite{deng2021depth} & 37.4 (\textcolor{green}{-4.2})& 3.0 (\textcolor{green}{-0.9})& 14.4 (\textcolor{green}{-2.2})& 27.4 (\textcolor{green}{-0.8})& 5.4 (\textcolor{green}{-1.0})& 13.6 (\textcolor{green}{-0.9})\\
    URF~\cite{rematas2021urban} & 46.4 (\textcolor{red}{3.0})& 4.5 (\textcolor{red}{0.3})& 18.4 (\textcolor{red}{1.6})& 28.3 (\textcolor{red}{0.1})& 5.3 (\textcolor{green}{-0.1})& 13.1 (\textcolor{red}{0.2})\\
    Ours& 32.0 (\textcolor{green}{-2.1})& 2.3 (\textcolor{green}{-2.5})& 9.0 (\textcolor{green}{-3.9})& 30.8 (\textcolor{green}{-2.1})& 5.1 (\textcolor{green}{-2.0})& 12.1 (\textcolor{green}{-2.3})\\
    \bottomrule
    \end{tabular}
    }
	\caption{Ablation study of volume rendering for active sensing.}
	\label{tab:supp_ablate_vol_render}
\end{table}
\begin{table}[t]
    \setlength{\tabcolsep}{4pt}
    \renewcommand{\arraystretch}{1.2}
	\centering
	\resizebox{0.95\columnwidth}{!}{
    \begin{tabular}{l|ccc|ccc|ccc}
    \toprule
    & \multicolumn{3}{c|}{TownClean} & \multicolumn{3}{c|}{TownReal} & \multicolumn{3}{c}{Waymo NVS} \\
    Method & Rec@5 $\uparrow$ & RE $\downarrow$ & TE $\downarrow$ &  Rec@5 $\uparrow$ & RE $\downarrow$ & TE $\downarrow$ & Rec@2 $\uparrow$ & RE $\downarrow$ &  TE $\downarrow$ \\
    \midrule
    i-NGP~\cite{muller2022instant} + L2 & 40.6 & 0.2 & 6.2 & 39.6 & 0.2 & 6.7 & 26.5 & 0.1 & 3.2\\
    i-NGP~\cite{muller2022instant} & 70.3 & 0.1 & 4.2 & 76.0 & 0.1 & 4.2 & 60.2 & 0.1 & 1.9\\
    DS-NeRF~\cite{deng2021depth} & 58.3 & 0.2 & 5.1 & 56.2 & 0.2 & 5.1 & 42.3 & 0.1 & 2.4\\
    URF~\cite{rematas2021urban} & 61.5 & 0.2 & 5.0 & 59.9 & 0.1 & 4.7 & 32.1 & 0.1 & 2.7\\
    LiDARsim~\cite{manivasagam2020lidarsim} & \textbf{82.8} & \textbf{0.1} & \textbf{3.4} & \underline{79.2} & \textbf{0.1} & \textbf{3.4} & \underline{62.8} & \textbf{0.1} & \underline{1.8}\\
    Ours & \underline{80.2} & \underline{0.1} & \underline{3.7} & \textbf{85.9} & \underline{0.1} & \textbf{3.4} & \textbf{71.9} & \textbf{0.1} & \textbf{1.7}\\
    \bottomrule
    \end{tabular}
     }
	\caption{Point cloud registration results on three datasets.}
	\label{tab:supp_registration}
\end{table}

\begin{table}[t]
\setlength{\tabcolsep}{4pt}
\renewcommand{\arraystretch}{1.2}
\centering
\resizebox{0.8\columnwidth}{!}{
\begin{tabular}{l|ccc|ccc}
\toprule
& \multicolumn{3}{c|}{Vehicle} & \multicolumn{3}{c}{Background} \\
Method & Recall $\uparrow$ & Precision $\uparrow$ & IoU $\uparrow$ & Recall $\uparrow$ & Precision $\uparrow$ & IoU $\uparrow$ \\
\midrule
i-NGP~\cite{muller2022instant} + L2 & 68.4 & \textbf{90.2} & 64.1 & \textbf{99.3} & 96.3 & 95.6\\
i-NGP~\cite{muller2022instant} & \underline{93.2} & 85.9 & \underline{80.9} & 98.3 & \underline{99.2} & \underline{97.6}\\
DS-NeRF~\cite{deng2021depth} & 90.7 & \underline{87.1} & 80.2 & \textbf{98.5} & 98.9 & 97.4\\
URF~\cite{rematas2021urban} & 87.8 & 81.7 & 73.7 & 98.0 & 98.4 & 96.5\\
Lidarsim~\cite{manivasagam2020lidarsim} & 90.5 & 70.5 & 65.9 & 94.9 & 99.0 & 94.0\\
Ours & \textbf{95.9} & 87.0 & \textbf{83.9} & 98.3 & \textbf{99.5} & \textbf{97.8}\\
\bottomrule
\end{tabular}
}
\vspace{-3mm}
\caption{Semantic segmentation results on \textit{Waymo NVS} dataset.}
\label{tab:supp_sem_seg}
\end{table}

We perform further experiments to evaluate an additional baseline method, denoted as \textit{i-NGP\cite{muller2022instant} + L2}, which optimizes the range estimation through L2 loss~\cite{deng2021depth,rematas2021urban}. The comprehensive results of our experimentation are presented in \cref{tab:supp_main}, \cref{tab:supp_ablate_vol_render}, \cref{tab:supp_registration}, and \cref{tab:supp_sem_seg}. Our findings reveal that the L2 loss performs inferior to its L1 loss counterpart (\ie i-NGP~\cite{muller2022instant}). However, replacing the standard volume rendering with the proposed formulation for active sensors, still leads to improved performance, as demonstrated in \cref{tab:supp_ablate_vol_render}.

\begin{figure*}[t]
\centering
\includegraphics[width=1.0\textwidth]{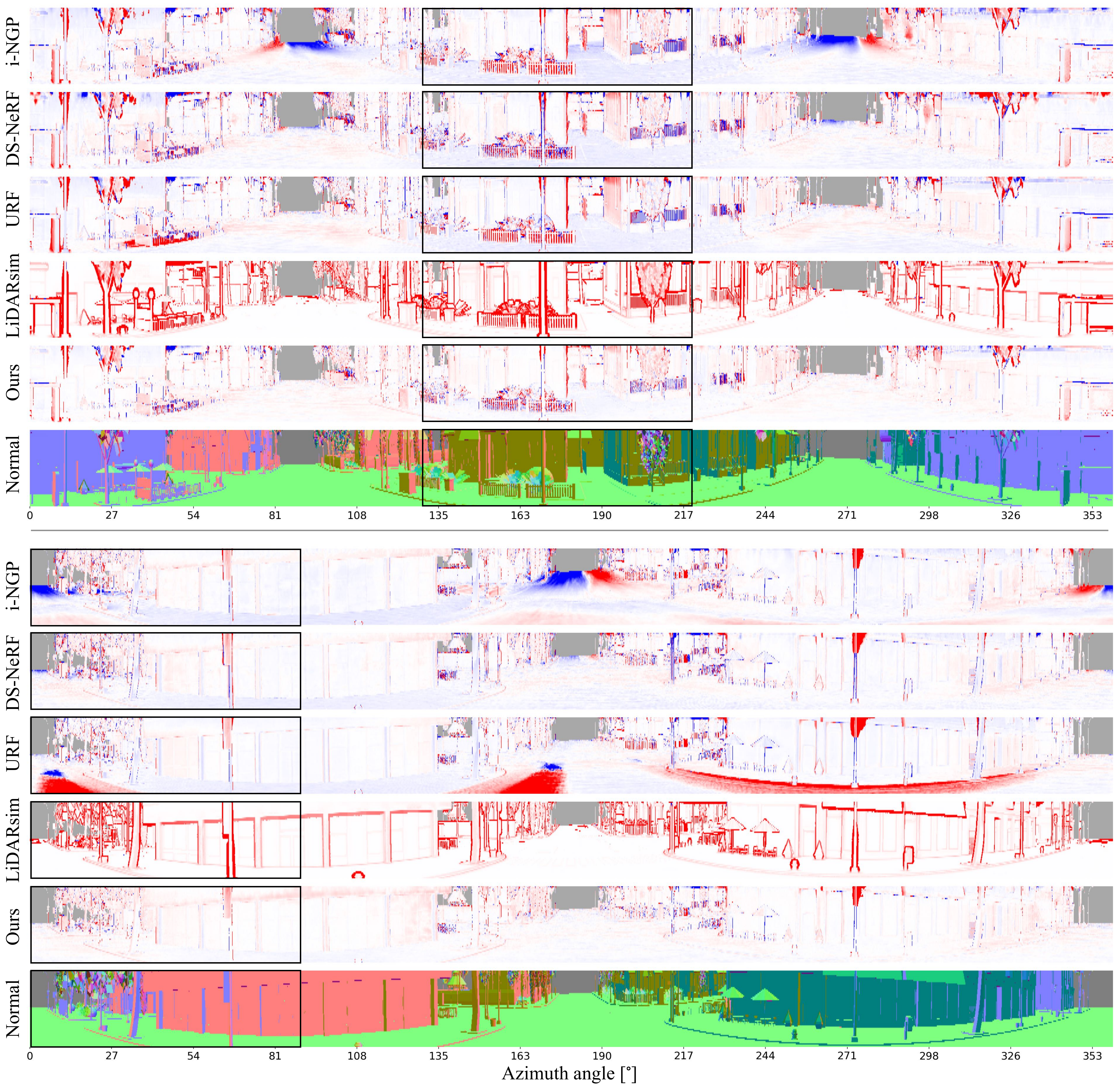}
\vspace{-3mm}
\caption{Qualitative results of first range estimation on \textit{TownClean} dataset.}
\label{fig:supp_townclean}
\vspace{-3mm}
\end{figure*}

\begin{figure*}[t]
\centering
\includegraphics[width=1.0\textwidth]{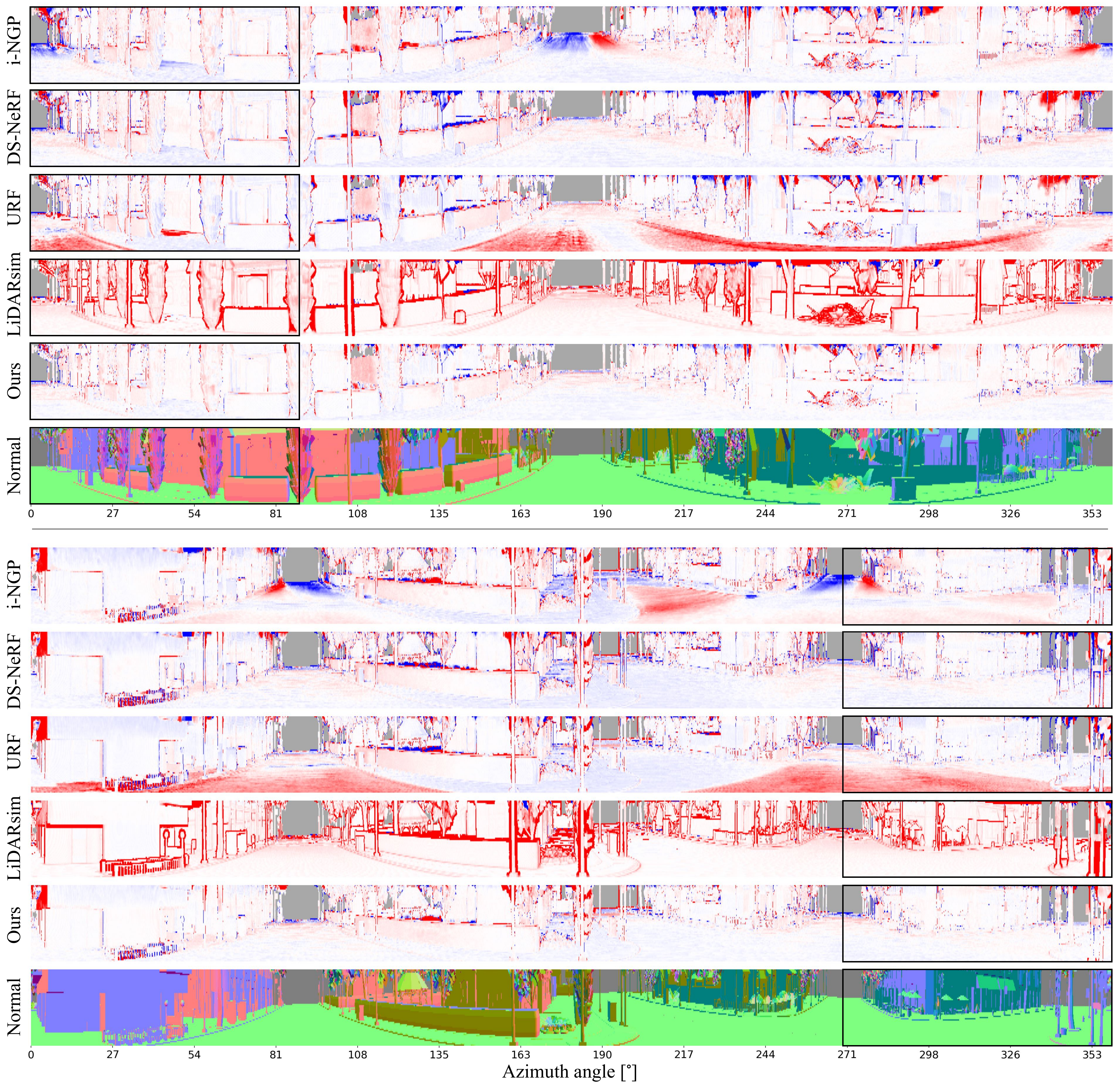}
\vspace{-3mm}
\caption{Qualitative results of first range estimation on \textit{TownReal} dataset.}
\label{fig:supp_townreal}
\vspace{-3mm}
\end{figure*}

\begin{figure*}[t]
\centering
\includegraphics[width=1.0\textwidth]{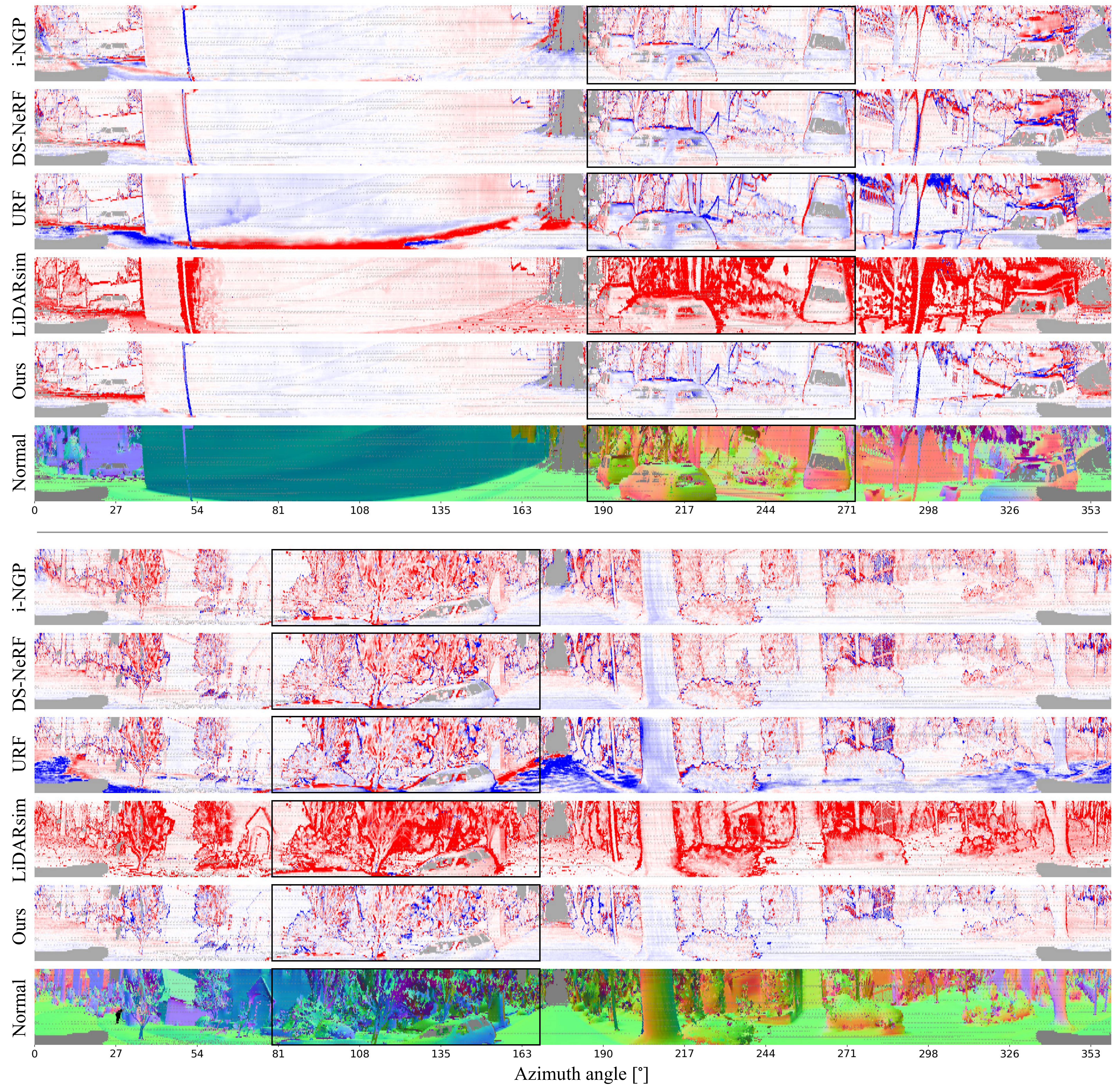}
\vspace{-3mm}
\caption{Qualitative results of first range estimation on \textit{Waymo NVS} dataset.}
\label{fig:supp_waymo_nvs}
\vspace{-3mm}
\end{figure*}

\begin{figure*}[t]
\centering
\includegraphics[width=1.0\textwidth]{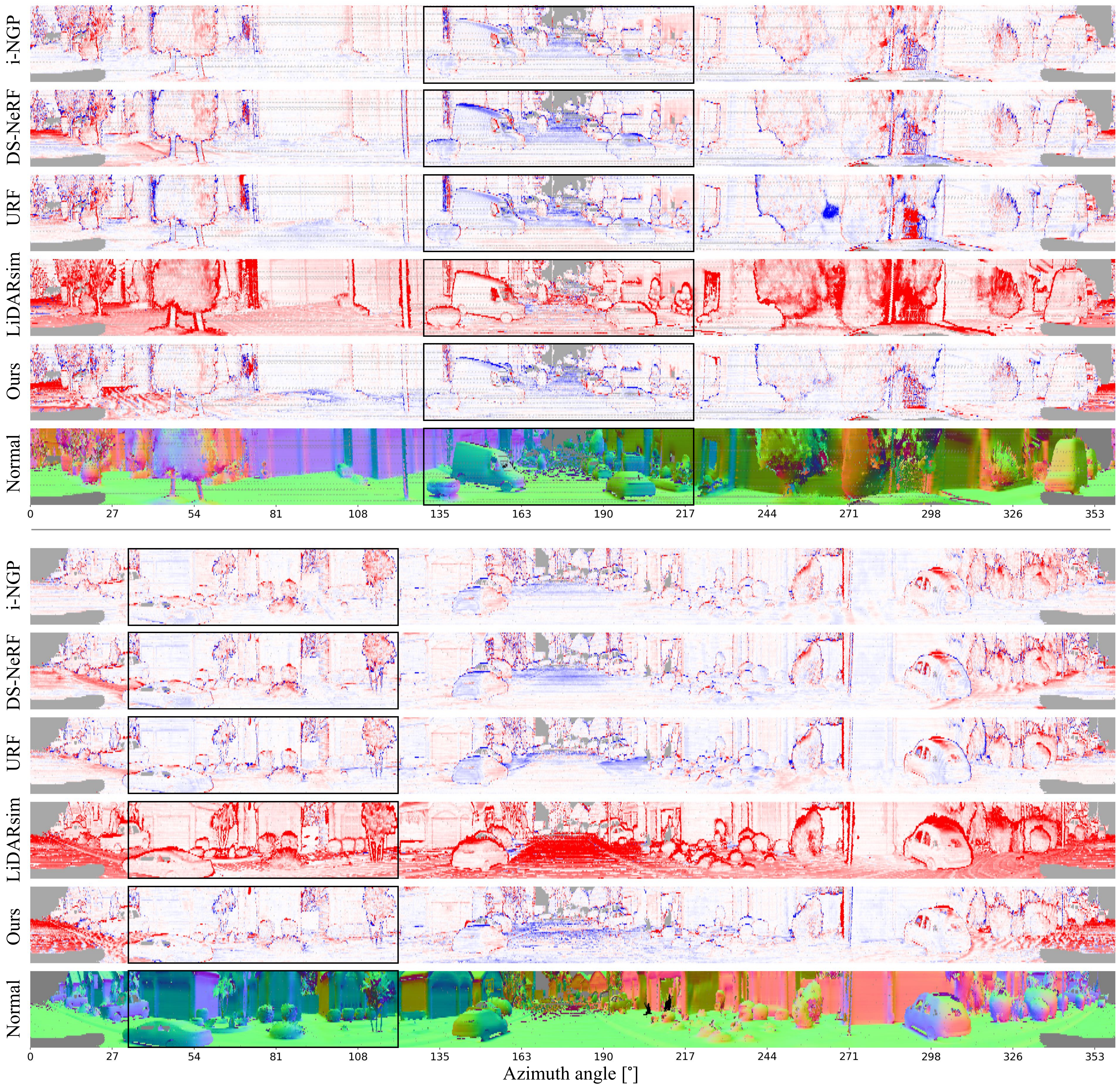}
\vspace{-3mm}
\caption{Qualitative results of first range estimation on \textit{Waymo Interp.} dataset.}
\label{fig:supp_waymo_interp}
\vspace{-3mm}
\end{figure*}

\paragraph{Qualitative results}
We show additional qualitative results in~\cref{fig:supp_townclean},~\cref{fig:supp_townreal},~\cref{fig:supp_waymo_nvs}, and ~\cref{fig:supp_waymo_interp}. We sample the middle frame of each dataset and present the first range errors in range-view projection. 

\clearpage


\end{document}